\documentclass{article}

\usepackage{arxiv}
\usepackage{amsmath, amssymb, amsthm}
\usepackage{algorithm}
\usepackage{algpseudocode}
\usepackage{geometry}
\usepackage{booktabs}
\usepackage{siunitx}
\usepackage{multirow}
\usepackage[table]{xcolor}
\usepackage{subfigure}
\usepackage{array} % For fixed-width columns
\usepackage{multirow} % For multirow functionality
\usepackage[utf8]{inputenc} % allow utf-8 input
\usepackage[T1]{fontenc}    % use 8-bit T1 fonts
\usepackage{url}            % simple URL typesetting
\usepackage{booktabs}       % professional-quality tables
\usepackage{amsfonts}       % blackboard math symbols
\usepackage{nicefrac}       % compact symbols for 1/2, etc.
\usepackage{microtype}      % microtypography
\usepackage{lipsum}		% Can be removed after putting your text content
\usepackage{graphicx}

\usepackage{doi}

\title{Mixed-Precision Conjugate Gradient Solvers with RL-Driven Precision Tuning\thanks{This work is in progress and will be submitted soon.}}

\author{ \href{https://orcid.org/0000-0003-1778-393X}{\includegraphics[scale=0.06]{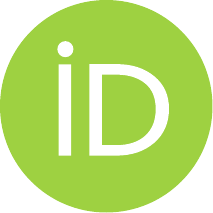}\hspace{1mm}Xinye Chen} \\
Sorbonne Université, CNRS, LIP6 \\
Paris, France \\
\texttt{xinye.chen@lip6.fr} 
}

\renewcommand{\shorttitle}{Mixed-Precision Conjugate Gradient Solvers with RL-Driven Precision Tuning}

\hypersetup{
pdftitle={Mixed-Precision Conjugate Gradient Solvers with RL-Driven Precision Tuning},
pdfsubject={q-bio.NC, q-bio.QM},
pdfauthor={Xinye Chen},
pdfkeywords={mixed precision algorithm, reinforcement learning, precision tuning, iterative linear solvers},
}

\begin{document}
\maketitle

\begin{abstract}
This paper presents a novel reinforcement learning (RL) framework for dynamically optimizing numerical precision in the conjugate gradient (CG) method. By modeling precision selection as a Markov Decision Process (MDP), we employ Q-learning to adaptively assign precision levels to key operations, striking an optimal balance between computational efficiency and numerical accuracy, while ensuring stability through double-precision scalar computations and residual computing. In practice, the algorithm is trained on a set of data and subsequently performs inference for precision selection on out-of-sample data, without requiring re-analysis or retraining for new datasets. This enables the method to adapt seamlessly to new problem instances without the  computational overhead of recalibration. Our empirical results demonstrate the effectiveness of our RL framework for mixed-precision CG solver, marking the first application of RL for precision selections for mixed-precision numerical algorithms. The empirical results highlight the approach's practical values, offering valuable insights into its extension to other iterative solvers and paving the way for AI-driven advancements in scientific computing.
\end{abstract}


\section{Introduction}
\label{sec:introduction}

Solving large-scale linear systems of the form $Ax = b$, where $A \in \mathbb{R}^{n \times n}$ is sparse and symmetric positive definite (SPD), is fundamental to computational science. These systems are central to applications ranging from finite element methods in structural engineering \cite{zienkiewicz2005finite} to gradient-based optimization in machine learning \cite{bottou2018optimization} and high-fidelity simulations in climate modeling \cite{dennis2010computational}. The scale of such systems, often involving millions or billions of unknowns, makes direct solvers computationally infeasible due to their quadratic or cubic complexity in memory and time \cite{davis2006direct}. Iterative solvers, such as the conjugate gradient (CG) method \cite{hestenes1952methods}, are widely adopted for their ability to exploit matrix sparsity and scale efficiently to large problems \cite{saad2003iterative}. 

Double-precision arithmetic ($\texttt{fp64}$) has traditionally been the standard for iterative solvers, ensuring high accuracy and robustness at the cost of significant computational and memory overhead \cite{demmel1997applied}. The emergence of modern computing architectures, such as graphics processing units (GPUs) and tensor processing units (TPUs), has fueled interest in lower-precision formats like single precision ($\texttt{fp32}$) and half precision ($\texttt{fp16}$), which reduce memory bandwidth requirements and accelerate computations \cite{markidis2018nvidia}. Low-precision arithmetic is favored in computational applications due to its faster arithmetic, reduced communication overhead, and lower energy consumption. However, the indiscriminate use of low-precision arithmetic can lead to numerical instability, delayed convergence, or divergence, particularly in ill-conditioned systems where small errors may accumulate catastrophically \cite{higham2019mixed, carson2021mixed}. Therefore,  the efficacy of numerical solvers is critically dependent on the selection of numerical precision, which dictates the balance among computational efficiency, memory requirements, and numerical stability. Mixed-precision strategies, which assign different precision levels to distinct operations based on their numerical sensitivity, offer a promising approach to balance these competing objectives \cite{baboulin2009accelerating}. Designing effective mixed-precision algorithms, however, remains challenging, as static or heuristic-based methods often fail to adapt to the dynamic, problem-specific behavior of iterative solvers. % see \cite{abdelfattah2021survey, osti_1814677} for survey and references therein.

Traditional precision tuning tools, such as Precimonious \cite{rubio2013precimonious} and PROMISE \cite{graillat2018promise}, have sought to address this challenge.  Both of them requires to run the program numerous times with the delta debugging algorithm \cite{10.1007/3-540-48166-4_16}. However, both of them lack an evaluation of generalization capability on new data, making it uncertain whether the original precision settings are applicable to unseen data, and cannot adapt to the runtime behavior of iterative solvers, where numerical properties evolve across iterations. These limitations underscore the need for a precision tuning framework that is both adaptive and computationally efficient, capable of optimizing performance across dynamic and diverse computational scenarios.

Reinforcement learning (RL), a machine learning paradigm rooted in sequential decision-making, provides a transformative solution to these challenges \cite{sutton2018reinforcement}. By modeling precision selection as a Markov Decision Process (MDP), RL enables the development of adaptive policies that dynamically optimize precision assignments based on the solver's evolving state, such as residual norms, convergence rates, or computational costs. Unlike PROMISE, which incurs significant computational overhead, RL leverages lightweight, model-free algorithms to learn optimal strategies through iterative interaction with the computational environment, minimizing runtime costs while maximizing a reward function that balances accuracy and efficiency \cite{mnih2015human}. In contrast to Precimonious and PROMISE, RL's dynamic adaptability allows it to respond to the unique numerical behavior of each iteration and problem instance. Moreover, RL's ability to generalize to out-of-sample data overcomes their context-specific limitations, enabling robust performance across a wide range of linear systems. By overcoming the limitations of traditional tools, our RL-based approach paves the way for a new generation of solvers that are faster, more efficient, and capable of tailoring their behavior to the unique characteristics of each problem.

In this work, we introduce the first application of RL to precision tuning and selection, focusing on the mixed-precision iterative solver{---}the CG method. We employ Q-learning, a robust and intuitive RL algorithm \cite{watkins1992q}, to dynamically control the precision of four key operations: matrix-vector multiplications, preconditioner applications, inner products, and vector updates. To ensure numerical stability, scalar operations and residual computing are maintained in $\texttt{fp64}$, while other operations are assigned to low precisions (e.g., \texttt{fp1}, \texttt{bf16}, \texttt{tf32}, and \texttt{fp32}, see the table~\ref{tab:unitroundoff} for detail) based on the learned policy. Our contributions include: (1) a comprehensive RL-based methodology for mixed-precision CG, with a detailed MDP formulation and Q-learning modeling and implementation; and (2) extensive numerical experiments demonstrating significant computational savings without compromising accuracy, even for challenging SPD systems.

\begin{table}[h]
\caption{Floating point formats; $u$: unit roundoff. $x_{\min}$: smallest positive normalized floating-point number. $x_{\max}$:  largest floating-point number. $t$: number of binary digits in the significand (including the implicit leading bit), $e_{\min}$: exponent of $x_{\min}$, and $e_{\max}$: exponent of $x_{\max}$.} 
\label{tab:unitroundoff} 
\centering \setlength\tabcolsep{3pt}\scriptsize
\begin{tabular}{l l l l r r r} 
\hline\\[-1.5mm]
& \qquad $u$      & \quad $x_{\min}$  &    \quad $x_{\max}$  &  $t$    &  $e_{\min}$  & $e_{\max}$\\[1mm] \hline\\[-0.6mm]
quater precision (\texttt{q52}) &  $1.25 \times 10^{-1}$  &  $6.10 \times 10^{-5}$&   $5.73 \times 10^{4}$ &        -16  &    -14   &   15 \\
 bfloat16 (bf16) &   $3.91\times 10^{-3}$  &  $1.18\times 10^{-38}$  & $3.39\times 10^{38}$   & 8     &   -126 & 127 \\
 half precision (\texttt{fp16})  &  $4.88\times 10^{-4}$ &   $6.10 \times 10^{-5}$ &  $6.55\times 10^{4}$&   11   &   -14 &     15 \\
 single (\texttt{fp32})  &  $5.96\times 10^{-8}$  &  $1.18\times 10^{-38}$ &   $3.40\times 10^{38}$ &  24    &  -126 &    127 \\
 double (\texttt{fp64})  &  $1.11\times 10^{-16}$  &  $2.23\times 10^{-308}$ & $1.80\times 10^{308}$ &  53  &   -1022  &  1023 \\[1mm]
\hline %inserts single line 
\end{tabular}
\end{table}

This study marks a pioneering effort to bridge reinforcement learning and numerical linear algebra, offering a novel framework for intelligent, adaptive computational methods.  This paradigm shift has the potential to accelerate scientific discovery across disciplines, from enabling real-time climate simulations to optimizing large-scale machine learning models. We structure the remainder of this paper as follows. Section~\ref{sec:related} provides an overview of recent trends in mixed-precision algorithms and precision tuning. Section~\ref{sec:cg_method} introduces essential notions and the conjugate gradient (CG) method. Section~\ref{sec:rl_framework} details our reinforcement learning (RL)-based framework, including the Markov decision process (MDP) formulation and the Q-learning algorithm. Section~\ref{sec:experiments} presents the numerical results. Finally, Sections~\ref{sec:discussion} and~\ref{sec:conclusion} discuss the broader impact and future directions, inspiring researchers to harness AI-driven methods to redefine the boundaries of computational science.

\section{Related work}\label{sec:related}

Due to the reduced computational cost and energy consumption, the mixed-precision arithmetic has been widely used in numerical methods, particularly in large-scale linear algebra computations (see \cite{abdelfattah2021survey, osti_1814677} and references therein). A well-studied topic is mixed-precision iterative refinement, where the system matrix is often factored in low precision, while residual evaluation and solution updates are carried out in higher precision \cite{doi:10.1137/20M1316822, doi:10.1137/17M1122918}. This mixed-precision computing routine often bring substantial performance gains while preserving numerical stability for iterative refinement, particularly when integrated with preconditioned Krylov subspace solvers~\cite{carson2022mixed, anzt2020adaptive}.

Recent methods extend this strategy by incorporating subspace recycling within mixed-precision Krylov solvers to accelerate convergence across sequences of related linear systems~\cite{carson2019accelerating}. These are particularly effective in settings involving repeated solves, such as in time-dependent PDEs and optimization routines. Theoretical guarantees for convergence under mixed-precision arithmetic are derived in~\cite{higham2002accuracy, higham2019mixed}, which analyze backward error bounds and conditioning constraints, offering criteria for selecting suitable precision levels at different computational stages.

Precision-adaptive solvers have also been proposed to dynamically switch between arithmetic formats based on convergence behavior, heuristic thresholds, or online error estimation~\cite{anzt2020adaptive, anzt2021mixed}. While such approaches offer robustness and efficiency, their adaptivity is often guided by fixed rules or manually designed heuristics that may not generalize across different problem instances or solver configurations.

Outside numerical solvers, dynamic program analysis tools such as Precimonious~\cite{rubio2013precimonious} and PROMISE~\cite{graillat2018promise} provide static or semi-static type tuning by exploring the floating-point type space under correctness constraints. These methods typically operate offline and apply precision tuning to source code using search algorithms and error estimation. These tools highlight the importance of precision tuning frameworks that are both computationally efficient and adaptable to dynamic, data-dependent program behavior, particularly in the context of scientific computing workloads. While effective, they lack the runtime adaptability and granularity achievable compared with our method. 

In contrast, reinforcement learning provides a data-driven mechanism for learning precision selection strategies that dynamically adapt to the structure of the linear system, observed numerical properties, and solver behavior. An RL agent can learn precision policies by interacting with the solver environment, adjusting the numerical format at each step to optimize for convergence rate, computational cost, or energy efficiency. Unlike fixed-rule methods, RL-based strategies can generalize from training data and improve over time through experience, making them particularly well-suited for iterative solvers operating in dynamically changing or data-dependent regimes. Though there is lack study of using RL for mixed precision algorithms, existing research has verified the performance of using RL for bitwidth adaptation for integer quantization of neural networks such as \cite{10566890} and \cite{10.1145/3649329.3656231}.

RL-based precision control frameworks align with recent advance in performance-aware numerical methods. In science and engineering, Krylov subspace methods have been widely employed in solving large-scale least-squares problems and regularized regression models~\cite{luo2024neural, deroos2017krylov}. The applications of Krylove subspace methods typically exhibits variable conditioning, repeated matrix structures, and heterogeneous numerical behavior—properties that increase the need for adaptive, learnable precision control for both compute-intensive vs data-intensive workloads.

\section{The Conjugate Gradient Method}
\label{sec:cg_method}

The conjugate gradient (CG) method is a cornerstone of iterative methods in numerical linear algebra, designed to solve systems of linear equations $A x = b$, where $A \in \mathbb{R}^{n \times n}$ is a symmetric positive definite (SPD) matrix, $b \in \mathbb{R}^n$ is a given right-hand side vector, and $x \in \mathbb{R}^n$ is the vector to be solved. The SPD property ensures that $A$ is invertible and that the quadratic form $x^T A x$ is positive for all non-zero $x$, guaranteeing the existence and uniqueness of the solution. The CG method is Krylov subspace method, which is particularly well-suited for large, sparse systems, as it requires only matrix-vector products and exhibits favorable convergence properties.

The CG method constructs a sequence of approximate solutions $\{ x_k \}$, residuals $\{ r_k \}$, and search directions $\{ p_k \}$, aiming to minimizes the quadratic function $\phi(x) = \frac{1}{2} x^T A x - b^T x$. At the $k$-th iteration, the algorithm proceeds as follows:
\begin{itemize}
    \item \textbf{Solution update}: The approximate solution is updated according to
    \begin{equation}
    \label{eq:solution_update}
    x_{k+1} = x_k + \alpha_k p_k,
    \end{equation}
    where $x_k \in \mathbb{R}^n$ is the current iterate, $p_k \in \mathbb{R}^n$ is the search direction, and $\alpha_k \in \mathbb{R}$ is a step size chosen to minimize $\phi(x_k + \alpha p_k)$ along $p_k$.

    \item \textbf{Residual update}: The residual is updated as
    \begin{equation}
    \label{eq:residual_update}
    r_{k+1} = r_k - \alpha_k A p_k,
    \end{equation}
    where $r_k = b - A x_k$ is the residual at iteration $k$. This update ensures that $r_{k+1} = b - A x_{k+1}$.

    \item \textbf{Search direction update}: The new search direction is computed as
    \begin{equation}
    \label{eq:direction_update}
    p_{k+1} = z_{k+1} + \beta_k p_k,
    \end{equation}
    where $z_{k+1} = M^{-1} r_{k+1}$ is the preconditioned residual, $M \in \mathbb{R}^{n \times n}$ is a symmetric positive definite preconditioner, and $\beta_k \in \mathbb{R}$ is a scalar coefficient. In the absence of preconditioning, $M = I$, and thus $z_{k+1} = r_{k+1}$.
\end{itemize}

The step size $\alpha_k$ and coefficient $\beta_k$ are chosen to enforce conjugacy of the search directions with respect to the $A$-inner product (i.e., $p_i^T A p_j = 0$ for $i \neq j$) and orthogonality of the residuals (i.e., $r_i^T r_j = 0$ for $i \neq j$). Specifically, they are given by
\begin{equation}
\label{eq:alpha_beta}
\alpha_k = \frac{r_k^T z_k}{p_k^T A p_k}, \quad \beta_k = \frac{r_{k+1}^T z_{k+1}}{r_k^T z_k}.
\end{equation}

These choices ensure that the CG method generates a sequence of iterates that lie in the Krylov subspace
\[
\mathcal{K}_k(A, r_0) = \text{span} \{ r_0, A r_0, A^2 r_0, \dots, A^{k-1} r_0 \},
\]
where $r_0 = b - A x_0$ is the initial residual. In exact arithmetic, the CG method converges to the exact solution in at most $n$ iterations, as the Krylov subspace spans $\mathbb{R}^n$. In practice, convergence depends on the condition number of $A$, and preconditioning (via a suitable $M$) can significantly accelerate convergence by reducing the effective condition number.

The CG method's efficiency stems from its minimal storage requirements (only a few vectors need to be stored) and its reliance on sparse matrix-vector products. These properties, combined with its theoretical guarantees, make it a preferred method for solving large-scale SPD systems in scientific computing and related fields.

% \subsection{Mixed Precision Algorithms with RL strategy}

\section{RL Framework for Iterative Solvers}
\label{sec:rl_framework}

In this section, we formulate a RL framework to optimize precision selection in iterative solvers for $Ax = b$, with SPD $A$. The framework assigns precisions to $m$ operations per iteration, drawn from $\mathcal{P}$. 
In the following, we will formulate our mixed-precision algorithm with RL strategy for precision selection.  Let $\mathcal{P} = \{ p_1, \ldots, p_m \}$ denote floating-point precisions, with unit roundoff $u_j \approx 2^{-m_j}$ ($m_j$ significand bits) and cost $c(p_j) > 0$. For example, $u_{\texttt{fp64}} \approx 2^{-53}$, but $c(\texttt{fp64}) > c(\texttt{fp32})$. Mixed-precision assigns precisions to operations, leveraging error tolerances. We denote rounding as $\text{fl}_{p_j}(x) = x (1 + \delta_j)$, $|\delta_j| \leq u_j$, $\| \cdot \|$ as the norm operator, and we use $\text{fl}_{p_j}(\cdot)$ to denote computation for operation $(\cdot)$ in precision $p_j$. In CG, matrix-vector products tolerate lower precision, but scalars require high precision for stability \cite{demmel1997applied, higham2019mixed}.

We choose Q-learning for its simplicity, robustness, and suitability for discrete, finite Markov Decision Process (MDP), making it ideal for precision optimization where states and actions are well-defined. Q-learning effectively utilizes finite, discrete state and action spaces ($|\mathcal{S}| = b \cdot r$, $|\mathcal{A}_j| = |\mathcal{P}|$), which are well-suited for straightforward tabular representations, eliminates the need for complex function approximators or transition models, ensures convergence to optimal policies with adequate exploration \cite{watkins1992q}, and provides modularity through separate Q-tables per operation that scale linearly with $m$. In contrast, policy gradient methods are impractical due to the challenges of gradient computations for discrete precisions, and deep reinforcement learning adds unnecessary training complexity. Q-learning’s blend of simplicity and rigor makes it the preferred choice for our methodology.

RL models sequential decision-making via an MDP $(\mathcal{S}, \mathcal{A}, R, P, \gamma)$ where $\mathcal{S}$ indicates \emph{states} for encoding system conditions, $\mathcal{A}$ is referred to as \emph{actions} for defining decisions;  $R: \mathcal{S} \times \mathcal{A} \to \mathbb{R}$ is referred to as \emph{rewards} for guiding optimization. $P(s' | s, a)$ is referred to as Transitions for modeling state evolution; $\gamma \in [0, 1)$ is \emph{discount factor} for balancing short- and long-term goals. 

The optimal policy $\pi^*: \mathcal{S} \to \mathcal{A}$ maximizes:
\[
\mathbb{E} \left[ \sum_{t=0}^\infty \gamma^t R(s_t, a_t) \right].
\]

Below, we detail the methodology, breaking it into intuitive components and practical considerations.

\subsection{Q-learning Mechanics}
The solver’s dynamics are modeled as an MDP, snapshotting its state to guide precision choices:

\begin{itemize}
    \item \textbf{State Space} $\mathcal{S}$: The state space is crafted to reflect CG’s iterative nature. Iteration bins ($i_k$) track progress, enabling RL to adjust precisions as iterations advance—e.g., using $\texttt{fp16}$ early when residuals are large, and $\texttt{fp64}$ later for refinement. Residual bins ($j_k$) use logarithmic scaling to capture orders of magnitude, critical for numerical stability \cite{higham2019mixed}. The floor $\epsilon_{\text{min}}$ prevents singularities in $\log_{10} \rho_k$. Parameters $b$ and $r$ balance granularity (large values for fine control) and learning speed (small values for smaller Q-tables). For example, $b = 10$, $r = 10$ yields $|\mathcal{S}| = 100$, suitable for typical CG runs.

    The state space $\mathcal{S}$ encodes solver progress via iteration $k \in \{0, \ldots, T_{\text{max}}-1\}$ and normalized residual $\rho_k = \frac{\| r_k \|}{\| b \|}$. 

    We discretize into:
        \begin{itemize}
            \item $b$ iteration bins: $i_k = \min \left( \lfloor k / \lceil T_{\text{max}} / b \rceil \rfloor, b-1 \right)$, dividing iterations evenly.
            \item $r$ residual bins: $j_k = \min \left( \lfloor -\log_{10} \max ( \rho_k, \epsilon_{\text{min}} ) / \delta \rfloor, r-1 \right)$, with $\epsilon_{\text{min}} > 0$, $\delta = -\log_{10} \epsilon_{\text{min}} / r$.
        \end{itemize}
        Thus, $s_k = (i_k, j_k)$ and $|\mathcal{S}| = b \cdot r$.

    \item \textbf{Action Space} $\mathcal{A}$: Actions assign precisions to $m$ operations, here $m = 5$ for CG (matrix-vector product, preconditioner solve, residual update, two inner products). The action space $\mathcal{P}^m$ is finite—e.g., for $\mathcal{P} = \{ \texttt{fp16}, \texttt{fp32}, \texttt{fp64} \}$, $|\mathcal{P}^5| = 3^5 = 243$. To scale, we use separate policies per operation, reducing complexity to $5 \cdot 3 = 15$. This modular design allows easy extension to other solvers (e.g., GMRES) or additional operations.
    
        An action is $a = (p_1, \ldots, p_m) \in \mathcal{P}^m$, assigning precisions to operations (e.g., matrix-vector product).

    \item \textbf{Reward Function}: The reward function is pivotal to our reinforcement learning (RL) approach, steering the Q-learning algorithm to optimize precision assignments in the conjugate gradient (CG) method. It balances numerical accuracy, computational efficiency, and convergence speed through three carefully designed components, integrated into a weighted sum that guides the RL agent’s policy.

The reward function is:
\[
R(s, a) = \omega_1 \min \left( -\log_{10} \rho', -\log_{10} \epsilon_{\text{min}} \right) - \omega_2 \sum_{j=1}^m c(p_j) + \omega_3 \mathbb{I} \left( \rho' < \tau \right),
\]
where $s$ and $a$ are the state and action. Parameters $\omega_1$, $\omega_2$, and $\omega_3$ are tuned to prioritize accuracy (high $\omega_1$, e.g., for ill-conditioned systems), efficiency (high $\omega_2$, e.g., for sparse systems), or fast convergence (high $\omega_3$).

The \emph{accuracy term}, $\omega_1 \min \left( -\log_{10} \rho', -\log_{10} \epsilon_{\text{min}} \right)$, rewards small relative residuals, $\rho' = \| r' \| / \| b \|$, with $\omega_1 > 0$ and $\epsilon_{\text{min}}$ ensuring bounded rewards. The \emph{cost term}, $-\omega_2 \sum_{j=1}^m c(p_j)$, penalizes high-precision operations, where $\omega_2 > 0$ and $c(p_j)$ is the cost of the $j$-th operation’s precision $p_j$, promoting efficiency. The \emph{convergence bonus}, $\omega_3 \mathbb{I} \left( \rho' < \tau \right)$, with $\omega_3 > 0$, rewards residuals below a tolerance $\tau$, accelerating convergence. In practice, we may consider $\omega_1=1.0, \omega_2=0.1, \omega_3=10.0$  (used in the following simulations).

    \item \textbf{Transitions}:
        Deterministic, $s' = \text{discretize}(k+1, \rho')$, as CG updates are fixed for given precisions.

    \item \textbf{Discount Factor}:
        $\gamma \in [0, 1)$, prioritizing accuracy over long-term costs.
\end{itemize}

Q-learning updates Q-tables $Q_j: \mathcal{S} \times \mathcal{P} \to \mathbb{R}$ for each operation $j$:
\[
Q_j(s_k, p_j) \leftarrow Q_j(s_k, p_j) + \alpha \left( R(s_k, a_k) + \gamma \max_{p \in \mathcal{P}} Q_j(s_{k+1}, p) - Q_j(s_k, p_j) \right),
\]
where $\alpha \in (0, 1]$ is the learning rate, and $a_k = (p_1, \ldots, p_m)$. Assuming no prior knowledge, we initialize $Q_j(s, p) = 0$, and train over $E$ episodes, each running CG until $\rho_k < \tau$ or $k = T_{\text{max}}$. Separate Q-tables reduce dimensionality from $|\mathcal{P}|^m$ to $m |\mathcal{P}|$, enhancing scalability. Q-learning’s model-free nature avoids explicit transition models, relying on solver feedback, making it robust to matrix variations \cite{watkins1992q}. Inference uses $p_j = \arg\max Q_j(s_k, p)$, applying learned policies efficiently. % —e.g., for $m = 5$, $|\mathcal{P}| = 3$, we manage 15 tables instead of 243

% \subsection{Exploration Strategy}
We employ an \(\epsilon\)-greedy policy, selecting random precisions with probability $\epsilon$, or the best-known ($\arg\max Q_j$) otherwise. The exploration rate decays as:
\[
\epsilon = \epsilon_0 \left( 1 - \frac{e}{E} \right),
\]

where $\epsilon_0 \in (0, 1]$, $e$ is the episode, and $E$ is the total episodes. Early high $\epsilon$ (e.g., $\epsilon_0 = 0.9$) tests diverse precisions, uncovering efficient schedules. Later low $\epsilon$ exploits learned policies, refining performance. This balance enable Q-learning to explore the action space (e.g., trying $\texttt{fp16}$ for matrix-vector products) while converging to optimal choices (e.g., $\texttt{fp64}$ near $\tau$).

\subsection{Precision Optimization in CG}
\label{sec:cg_application}

We apply the RL framework to the Conjugate Gradient (CG) method for solving $Ax = b$. 
Since the convergence of CG method is highly related to the conditioning of $A$, and the better conditioned A is, the faster gradient descent will converge. The preconditioner $M \approx A$ uses a fixed precision $p_{\text{fixed}} \in \mathcal{P}$ \cite{meijerink1977iterative}, set here to $p_{\text{fixed}} = \mathrm{fp32}$. The precision set is $\mathcal{P} = \{ \mathrm{bf16}, \mathrm{fp16}, \mathrm{tf32}, \mathrm{fp32}, \mathrm{fp64} \}$. We optimize the following operations associated with precisions selected by the RL agent:
\begin{enumerate}
    \item Matrix-vector product: $q_k = \text{fl}_{p_1}(A p_k)$.
    \item Preconditioner solve: $z_{k+1} = \text{fl}_{p_2}(M^{-1} r_{k+1})$.
    \item Inner product: $\nu_k = \text{fl}_{p_3}(p_k^T q_k)$.
    \item Inner product: $\sigma_{k+1} = \text{fl}_{p_4}(r_{k+1}^T z_{k+1})$.
\end{enumerate}
The RL action is $a_k = (p_1, p_2, p_3, p_4) \in \mathcal{P}^4$. The residual update, $r_{k+1} = \texttt{fp64}(r_k - \alpha_k q_k)$, is computed in full precision (fp64). The scalars $\alpha_k = \frac{\sigma_k}{\nu_k}$ and $\beta_k = \frac{\sigma_{k+1}}{\sigma_k}$ are computed in fp64 using mixed-precision inputs ($\sigma_k, \nu_k, \sigma_{k+1}$) to ensure stability \cite{demmel1997applied}. The state $s_k \in \mathcal{S}$ is defined based on the iteration $k$ and the residual norm ratio $\rho_k = \| r_k \| / \| b \|$, following Section \ref{sec:rl_framework}.

The proposed framework leverages reinforcement learning (RL) to optimize floating-point precision selection in the preconditioned conjugate gradient (CG) method for solving linear systems $Ax = b$, where $A$ is symmetric positive definite. Operating in two phases, the framework first conducts a training phase over $E$ episodes, where each episode initializes the CG iterates ($x_0 = 0$, $r_0 = b$, $z_0 = M^{-1} r_0$, $p_0 = z_0$, $\sigma_0 = r_0^T z_0$) and iteratively computes the state $s_k = \text{discretize}(k, \| r_k \|_2 / \| b \|_2)$, selects precisions $p_j$ for five key operations using an $\epsilon$-greedy policy, executes the CG iteration, calculates a reward balancing accuracy and computational cost, and updates the Q-value tables $Q_j$. In the inference phase, the trained Q-values guide precision selection to efficiently solve the system while logging the chosen precisions, ensuring a balance between numerical accuracy and computational efficiency. The detail can be referred to Algorithm~\ref{alg:rl_cg_training} and Algorithm~\ref{alg:rl_cg_inference}.

\begin{algorithm}
\caption{Training Phase for RL-Driven Precision Selection in CG}
\label{alg:rl_cg_training}
\begin{algorithmic}[1]
\State \textbf{Input}: Matrix $A$, vector $b$, precision set $\mathcal{P}$, preconditioner $M$, tolerance $\tau$, maximum iterations $T_{\text{max}}$, episodes $E$, initial exploration rate $\epsilon_0$, learning rate $\alpha$, discount factor $\gamma$, reward coefficients $\omega_1, \omega_2, \omega_3$, minimum residual threshold $\epsilon_{\text{min}}$
\State \textbf{Output}: Q-value tables $Q_j(s, p)$, for $j = 1, \ldots, 4$, $s \in \mathcal{S}$, $p \in \mathcal{P}$
\State Initialize $Q_j(s, p) = 0$, for $j = 1, \ldots, 4$, $s \in \mathcal{S}$, $p \in \mathcal{P}$
\For{episode $e = 1, \ldots, E$}
    \State Initialize $x_0 \gets 0$, $r_0 \gets b$, $z_0 \gets M^{-1} r_0$, $p_0 \gets z_0$
    \State Compute $\sigma_0 \gets r_0^T z_0$, $b_{\text{norm}} \gets \| b \|_2$
    \State Set exploration rate $\epsilon \gets \epsilon_0 (1 - e/E)$
    \For{iteration $k = 0, \ldots, T_{\text{max}}-1$}
        \State Compute state $s_k \gets \text{discretize}(k, \| r_k \|_2 / b_{\text{norm}})$
        \State Select precisions $p_j \gets \begin{cases} 
            \text{random } p \in \mathcal{P} & \text{if } \text{rand}() < \epsilon \\
            \arg\max_{p \in \mathcal{P}} Q_j(s_k, p) & \text{otherwise}
        \end{cases}$, for $j = 1, \ldots, 4$
        \State Compute $q_k \gets \text{fl}_{p_1}(A p_k)$ \Comment{Matrix-vector product}
        \State Compute $\nu_k \gets \text{fl}_{p_3}(p_k^T q_k)$ \Comment{Inner product for $\alpha_k$}
        \State Compute $\alpha_k \gets \sigma_k / \nu_k$ % \Comment{Using fp64}
        \State Update $x_{k+1} \gets x_k + \alpha_k p_k$
        \State Compute $r_{k+1} \gets r_k - \alpha_k q_k$ %\Comment{Residual update in fp64 precision}
        \If{$\| r_{k+1} \|_2 / b_{\text{norm}} < \tau$}
            \State Compute reward $R \gets \omega_1 \min \left( -\log_{10} (\| r_{k+1} \|_2 / b_{\text{norm}}), -\log_{10} \epsilon_{\text{min}} \right) + \omega_3 - \omega_2 \sum_{j=1}^5 c(p_j)$
            \State Update $Q_j(s_k, p_j) \gets Q_j(s_k, p_j) + \alpha \left( R - Q_j(s_k, p_j) \right)$, for $j = 1, \ldots, 4$
            \State \textbf{break}
        \EndIf
        \State Compute $z_{k+1} \gets \text{fl}_{p_2}(M^{-1} r_{k+1})$ \Comment{Preconditioner solve}
        \State Compute $\sigma_{k+1} \gets \text{fl}_{p_4}(r_{k+1}^T z_{k+1})$ \Comment{Inner product for $\beta_k$}
        \State Compute $\beta_k \gets \sigma_{k+1} / \sigma_k$ % \Comment{Using fp64}
        \State Update $p_{k+1} \gets z_{k+1} + \beta_k p_k$
        \State Compute reward $R \gets \omega_1 \min \left( -\log_{10} (\| r_{k+1} \|_2 / b_{\text{norm}}), -\log_{10} \epsilon_{\text{min}} \right) - \omega_2 \sum_{j=1}^5 c(p_j)$
        \State Compute next state $s_{k+1} \gets \text{discretize}(k+1, \| r_{k+1} \|_2 / b_{\text{norm}})$
        \State Update $Q_j(s_k, p_j) \gets Q_j(s_k, p_j) + \alpha \left( R + \gamma \max_{p \in \mathcal{P}} Q_j(s_{k+1}, p) - Q_j(s_k, p_j) \right)$, for $j = 1, \ldots, 5$
        \State Set $\sigma_k \gets \sigma_{k+1}$
    \EndFor
\EndFor
\State \textbf{Return} $Q_j(s, p)$, for $j = 1, \ldots, 4$
\end{algorithmic}
\end{algorithm}

\begin{algorithm}
\caption{Inference Phase for RL-Driven Precision Selection in CG}
\label{alg:rl_cg_inference}
\begin{algorithmic}[1]
\State \textbf{Input}: Matrix $A$, vector $b$, precision set $\mathcal{P}$, preconditioner $M$, tolerance $\tau$, maximum iterations $T_{\text{max}}$, Q-value tables $Q_j(s, p)$, for $j = 1, \ldots, 4$
\State \textbf{Output}: Solution $x$
\State Initialize $x_0 \gets 0$, $r_0 \gets b$, $z_0 \gets M^{-1} r_0$, $p_0 \gets z_0$
\State Compute $\sigma_0 \gets r_0^T z_0$, $b_{\text{norm}} \gets \| b \|_2$
\State Initialize $\text{log} \gets []$
\For{iteration $k = 0, \ldots, T_{\text{max}}-1$}
    \State Compute state $s_k \gets \text{discretize}(k, \| r_k \|_2 / b_{\text{norm}})$
    \State Select precisions $p_j \gets \arg\max_{p \in \mathcal{P}} Q_j(s_k, p)$, for $j = 1, \ldots, 4$
    % \State Append $(k, p_1, \ldots, p_4)$ to $\text{log}$
    \State Compute $q_k \gets \text{fl}_{p_1}(A p_k)$ \Comment{Matrix-vector product}
    \State Compute $\nu_k \gets \text{fl}_{p_3}(p_k^T q_k)$ \Comment{Inner product for $\alpha_k$}
    \State Compute $\alpha_k \gets \sigma_k / \nu_k$ %\Comment{Using fp64}
    \State Update $x_{k+1} \gets x_k + \alpha_k p_k$
    \State Compute $r_{k+1} \gets r_k - \alpha_k q_k$ % \Comment{Residual update in fp64 precision}
    \If{$\| r_{k+1} \|_2 / b_{\text{norm}} < \tau$}
        \State Append $(k+1, p_1, \ldots, p_4)$ to $\text{log}$
        \State \textbf{break}
    \EndIf
    \State Compute $z_{k+1} \gets \text{fl}_{p_2}(M^{-1} r_{k+1})$ \Comment{Preconditioner solve}
    \State Compute $\sigma_{k+1} \gets \text{fl}_{p_4}(r_{k+1}^T z_{k+1})$ \Comment{Inner product for $\beta_k$}
    \State Compute $\beta_k \gets \sigma_{k+1} / \sigma_k$ %\Comment{Using fp64}
    \State Update $p_{k+1} \gets z_{k+1} + \beta_k p_k$
    \State Set $\sigma_k \gets \sigma_{k+1}$
\EndFor
\State \textbf{Return} $x_{k+1}$
\end{algorithmic}
\end{algorithm}

\section{Experiments}
\label{sec:experiments}
In this experiment, the RL agent based on Q-learning dynamically selected precisions ($P = \{\mathrm{fp16}, \mathrm{bf16}, \mathrm{tf32}, \mathrm{fp32}, \mathrm{fp64}\}$) for four operations in CG\footnote{one can customize on their own}: matrix-vector product ($q_k = A p_k$), preconditioner solve ($z_{k+1} = M^{-1} r_{k+1}$), and two inner products ($\nu_k = p_k^T q_k$, $\sigma_{k+1} = r_{k+1}^T z_{k+1}$). The state space combined iteration index $k$ (10 bins over 1000 iterations) and log-scaled residual norm ratio $\rho_k = \| r_k \| / \| b \|$ (10 bins, minimum $10^{-16}$). Four Q-tables of size $100 \times 5$ were maintained, using an epsilon-greedy policy (decaying from 1.0 to 0.1) with learning rate $\alpha = 0.1$, discount factor $\gamma = 0.9$, and a reward balancing accuracy, convergence, and computational cost defined below:
\begin{enumerate}
    \item Cost setting $C_1$: $c(\mathrm{bf16})=0.6, c(\mathrm{fp16})=0.8, c(\mathrm{tf32})=0.8, c(\mathrm{fp32})=1.0, c(\mathrm{fp64}\})=2.0$
    \item Cost setting $C_2$: $c(\mathrm{bf16})=0.4, c(\mathrm{fp16})=0.5, c(\mathrm{tf32})=0.5, c(\mathrm{fp32})=1.5, c(\mathrm{fp64}\})=3.0$
\end{enumerate}

We perform training that involved 200 episodes per training matrix, updating Q-tables, executing the mixed-precision CG solver. We choose incomplete LU factorization computed in \texttt{fp32} (drop tolerance $10^{-4}$, fill factor 10) precision as preconditioner for CG method, and fixed the parameter of for preconditioners, though this setting does not necessarily work for all linear system, but it is interesting to see how RL works in practice for those bad-conditioned linear system. The fp64-CG solver, using double precision throughout with the same preconditioner, served as the reference.  

To assess the robustness and generalizability of our method under diverse conditions, particularly in data-scarce scenarios, we simulate experiments on both synthetic sparse linear systems and those derived from the Poisson problem. To emulate environments with limited training data and show effectiveness of our method where data is scarce, we deliberately restrict the training set size to $n_{\text{train}} = 10$. Besides, this setup reflects practical situations where only a small amount of data is available for model development. In contrast, the testing set is significantly larger ($n_{\text{test}} = 100$), enabling a tough evaluation of the model’s ability to generalize across a broader and more varied set of systems.

\begin{figure}[ht]
\centering
\subfigure[Sample 1 (Cost setting $C_1$)]{\includegraphics[width=6.3cm]{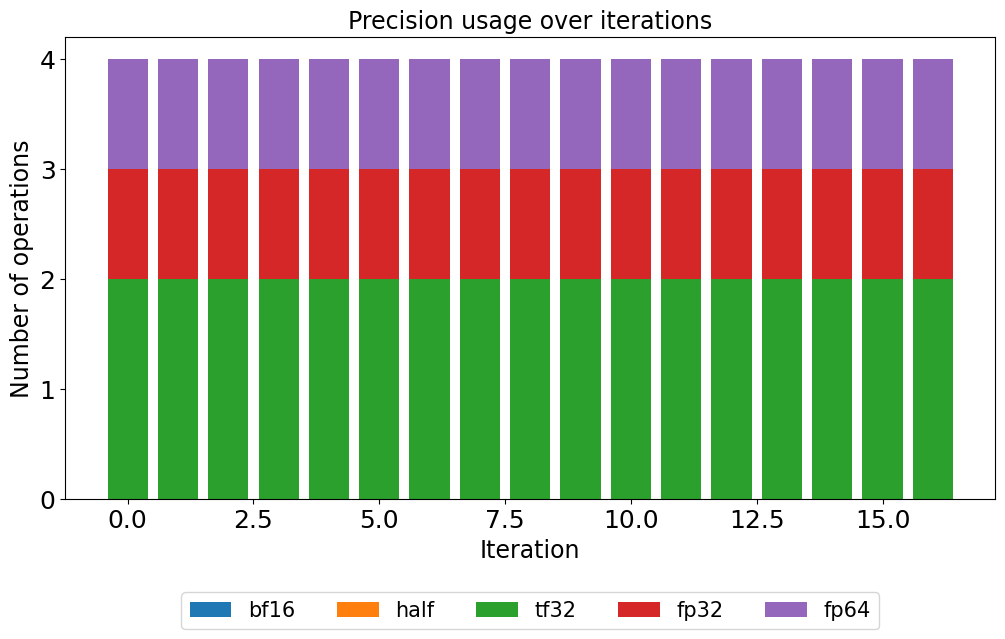}}
\subfigure[Sample 1 (Cost setting $C_2$)]{\includegraphics[width=6.3cm]{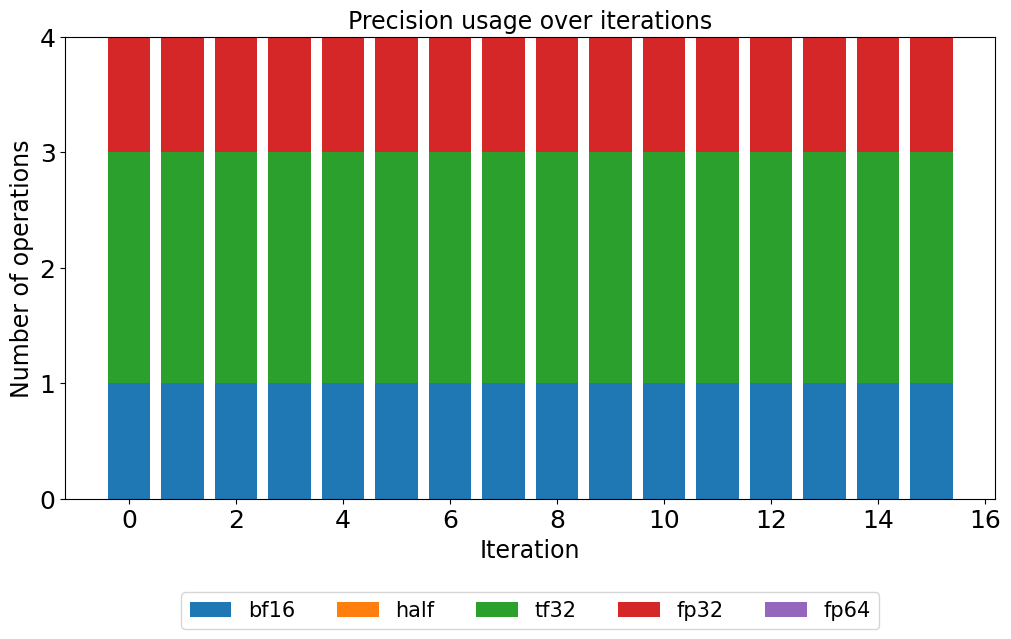}}\\
\subfigure[Sample 2 (Cost setting $C_1$)]{\includegraphics[width=6.3cm]{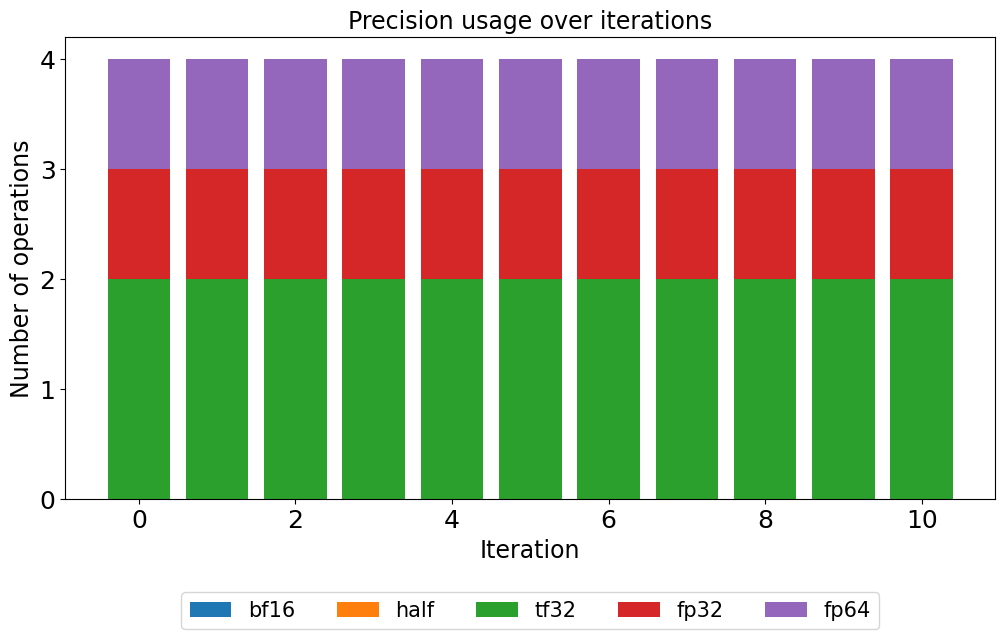}}
\subfigure[Sample 2 (Cost setting $C_2$)]{\includegraphics[width=6.3cm]{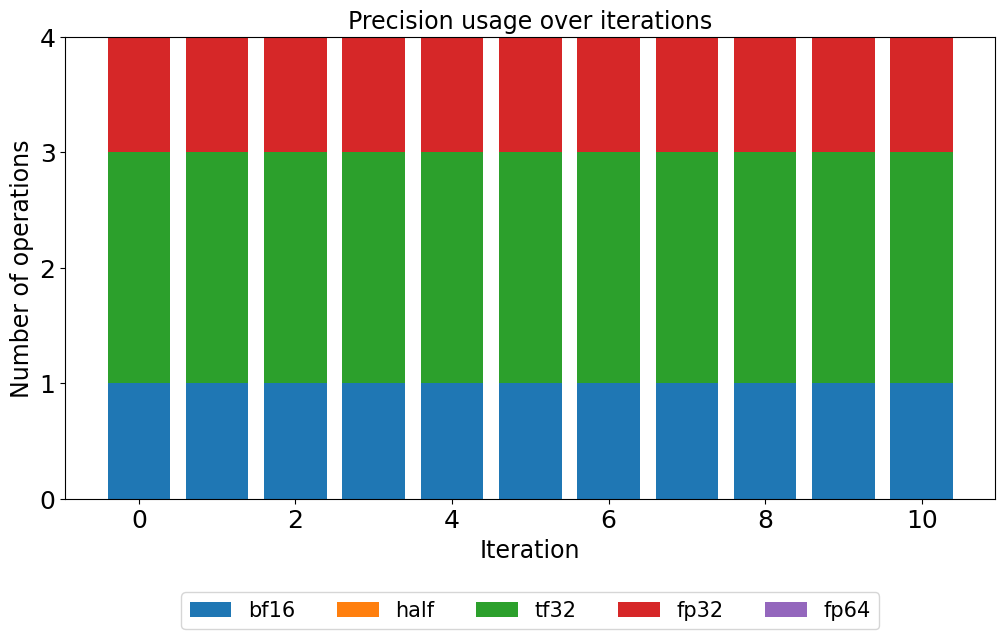}}\\
\subfigure[Sample 3 (Cost setting $C_1$)]{\includegraphics[width=6.3cm]{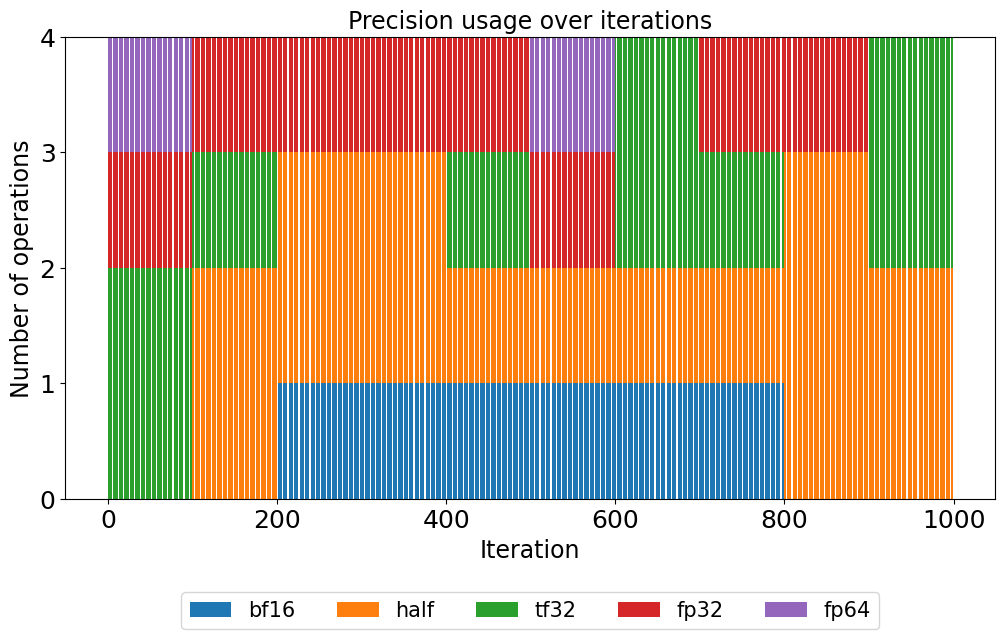}}
\subfigure[Sample 3 (Cost setting $C_2$)]{\includegraphics[width=6.3cm]{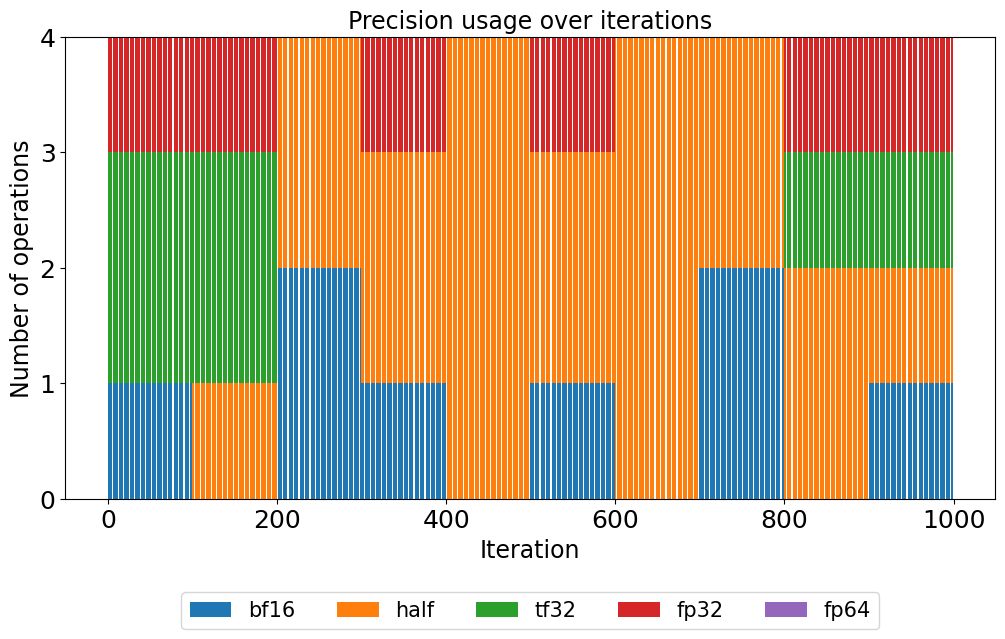}}\\
\caption{Precision selection for each iteration by RL.}\label{fig:prec_select1}
\end{figure}

\begin{figure}[ht]
\centering
\subfigure[Sample 1 (Cost setting $C_1$)]{\includegraphics[width=6.3cm]{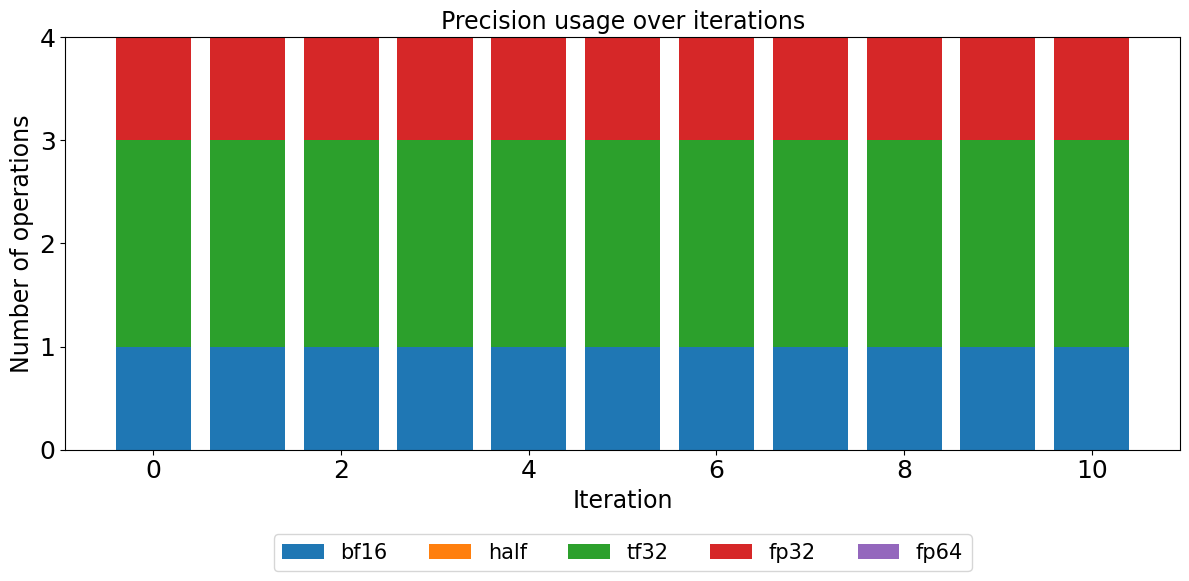}}
\subfigure[Sample 1 (Cost setting $C_2$)]{\includegraphics[width=6.3cm]{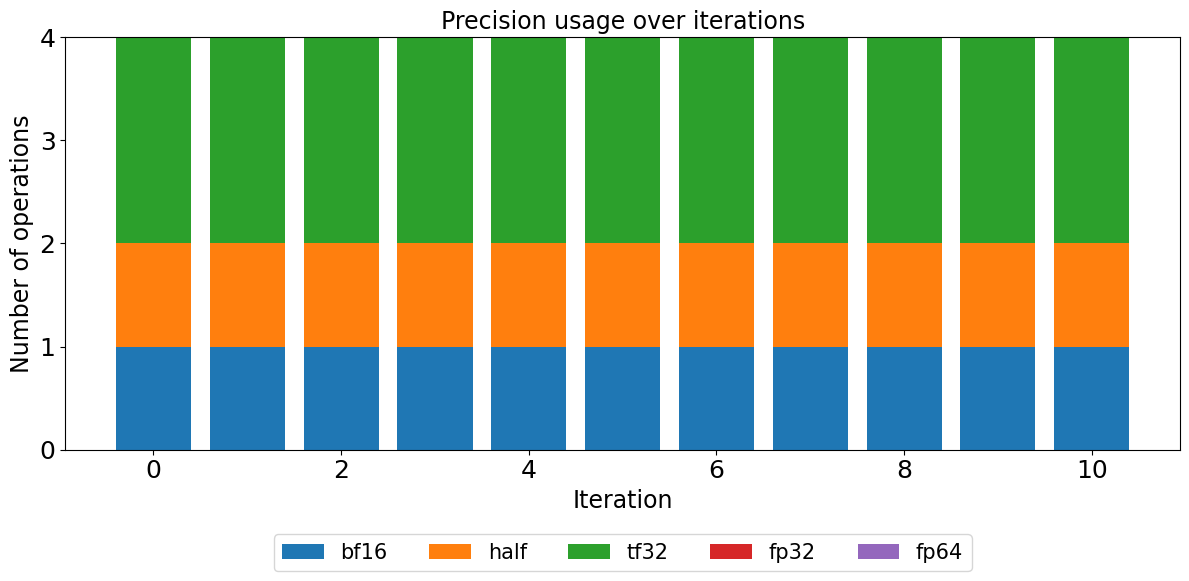}}\\
\subfigure[Sample 2 (Cost setting $C_1$)]{\includegraphics[width=6.3cm]{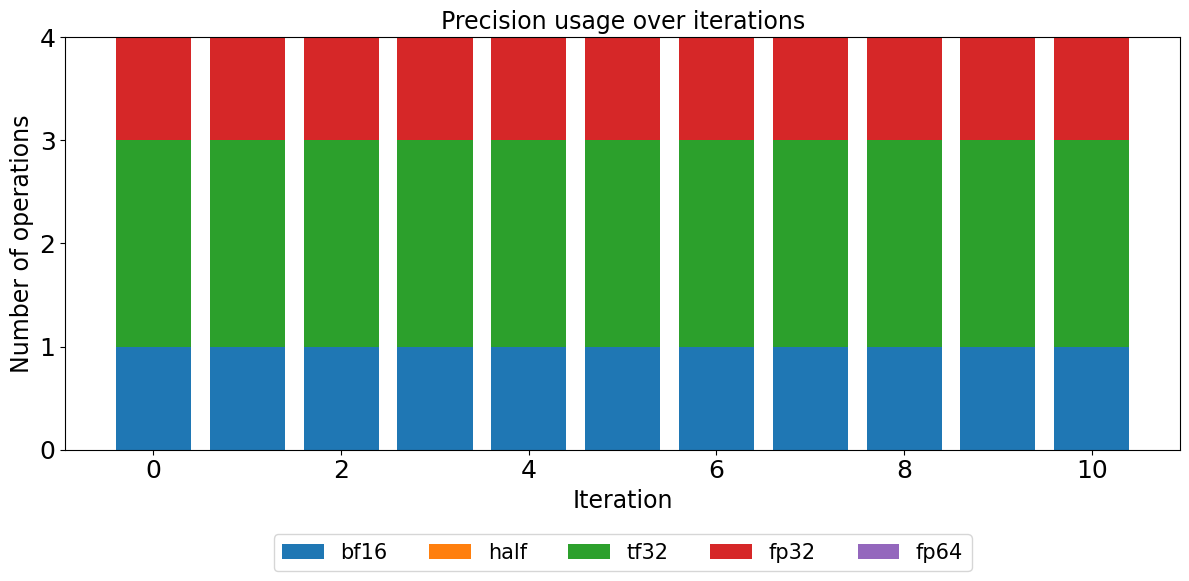}}
\subfigure[Sample 2 (Cost setting $C_2$)]{\includegraphics[width=6.3cm]{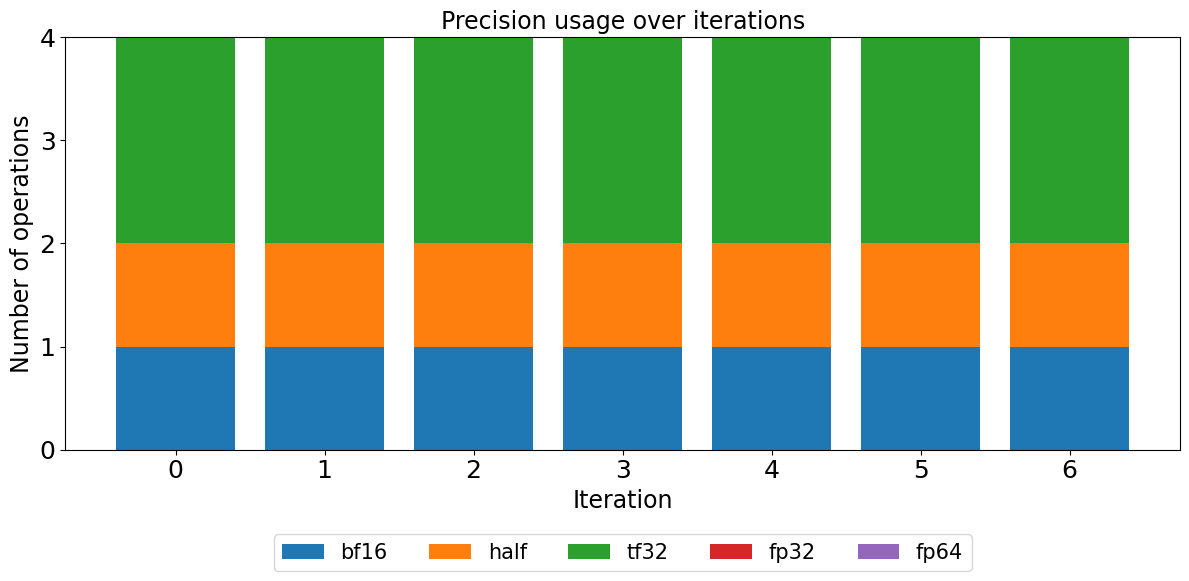}}\\
\subfigure[Sample 3 (Cost setting $C_1$)]{\includegraphics[width=6.3cm]{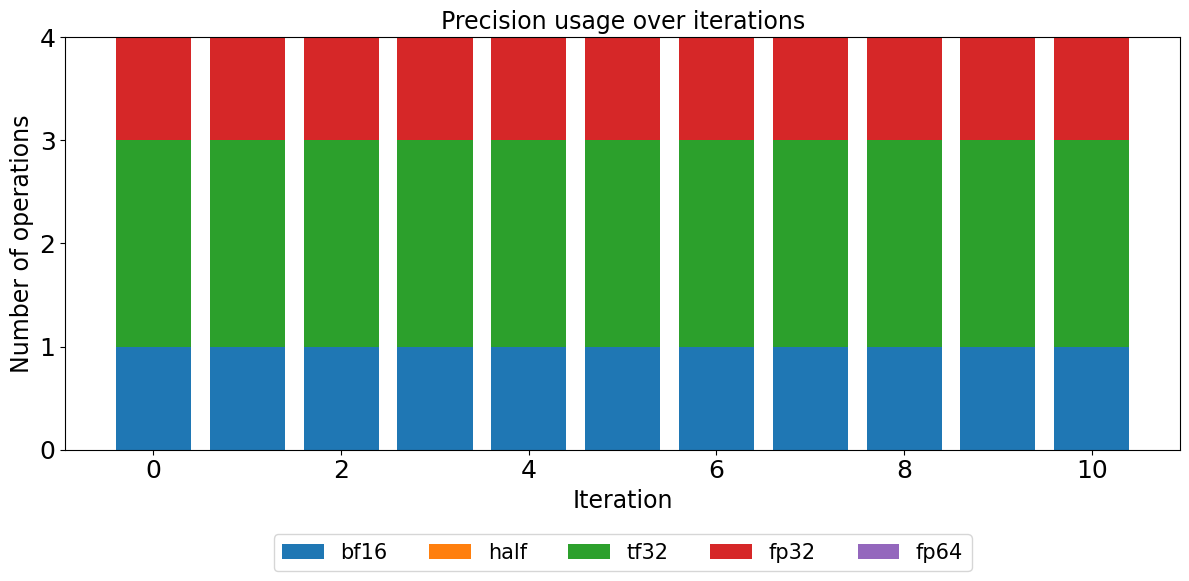}}
\subfigure[Sample 3 (Cost setting $C_2$)]{\includegraphics[width=6.3cm]{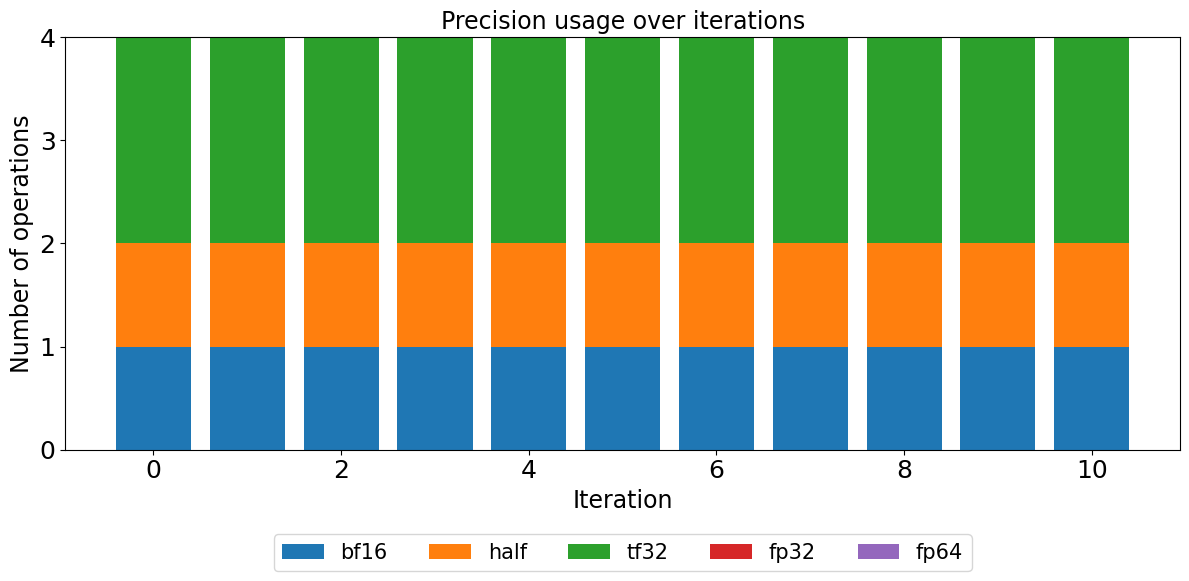}}\\
\caption{Precision selection for each iteration by RL.}\label{fig:prec_select2}
\end{figure}

Testing evaluated the trained RL agent on 100 test matrices, applying the greedy policy to select precisions for the mixed-precision CG solver and comparing against the fp64-CG solver. Both solvers terminated after 1,000 iterations, upon convergence ($\rho_k < 10^{-6}$, $k \geq 10$), or due to numerical instabilities. Performance was assessed via relative error ($\| x - x_{\text{true}} \| / \| x_{\text{true}} \|$, with $x_{\text{true}}$ from direct solve with LU decomposition and iteration count, recorded per matrix and averaged. Precision choices of the first three matrices for the two problems are depicted in \figurename~\ref{fig:prec_select1} and \figurename~\ref{fig:prec_select2}, respectively. Besides, the averaged number of the precision types used for each matrix for two problems are presented in \tablename~\ref{tab:average_precs}. The RL-based mixed-precision CG aimed to achieve comparable accuracy to the fp64-CG with reduced computational cost through lower precisions, while the fp64-CG provided a high-accuracy baseline. Throughout the remainder of this work, we refer to the RL-based mixed-precision CG solver as RL-CG for brevity.

Our experiments were performed on a Dell PowerEdge R750xa server equipped with 2 TB of memory, Intel Xeon Gold 6330 processors (56 cores, 112 threads, 2.00 GHz), and an NVIDIA A100 GPU (80 GB HBM2, PCIe). The computational framework leveraged \texttt{PyTorch} \cite{NEURIPS2019_9015} for reinforcement learning deployment and tensor computations, \texttt{SciPy} \cite{2020SciPy-NMeth} for manipulating sparse matrices, and \texttt{Pychop} \cite{carson2025} for low-precision emulation. All numerical results in the tables are rounded and shown with three significant digits.

\begin{table}[h]
\centering\footnotesize
\caption{Distribution of precision types used, shown as percentages of the total per setting.}
\label{tab:average_precs}
\begin{tabular}{llccccc}
\toprule
& & $\mathrm{fp16}$ & $\mathrm{bf16}$ & $\mathrm{tf32}$ & $\mathrm{fp32}$ & $\mathrm{fp64}$ \\
\midrule
\multirow{2}{*}{Synthetic sparse random dataset} 
& $C_1$ & 2.47\% & 6.19\% & 43.9\% & 23.5\% & 23.9\% \\
& $C_2$ & 22.3\% & 8.27\% & 40.7\% & 21.5\% & 7.19\% \\
\midrule
\multirow{2}{*}{2D Poisson PDE problems} 
& $C_1$ & 25.0\% & 0.00\% & 50.0\% & 25.0\% & 0.00\% \\
& $C_2$ & 25.0\% & 25.0\% & 50.0\% & 0.00\% & 0.00\% \\
\bottomrule
\end{tabular}
\end{table}

\subsection{Synthetic sparse random dataset}

We generated a synthetic dataset of linear systems \(\mathbf{A} \mathbf{x} = \mathbf{b}\) with sparse, positive definite matrices. A synthetic approach was chosen to provide controlled variability in matrix properties, such as sparsity and conditioning, enabling robust testing across diverse scenarios that mimic real-world numerical challenges while ensuring reproducibility.

The generation process constructs a sparse symmetric matrix \(\mathbf{A} \in \mathbb{R}^{5,000 \times 5,000}\), defined as \(\mathbf{A} = \mathbf{B} \mathbf{B}^T + \beta \mathbf{I}\), where \(\mathbf{B}\) is a sparse matrix with approximately 1\% of its elements being non-zero and randomly distributed, and \(\mathbf{I}\) denotes the identity matrix. The parameter \(\beta\), drawn from a uniform distribution \(\beta \sim \text{Uniform}(10^{-4}, 10^{-2})\), ensures that \(\mathbf{A}\) is positive definite. Non-zero entries of \(\mathbf{B}\) are specified by index pairs \((i_k, j_k)\), where row and column indices are sampled uniformly with replacement from \(\{0, 1, \ldots, 4999\}\), resulting in 5,000 such pairs. The corresponding values are drawn from a standard normal distribution, i.e., \(b_{i_k, j_k} \sim \mathcal{N}(0, 1)\). For out-of-distribution testing, the sparsity level is adjusted by a scaling factor sampled from \(\text{Uniform}(0.8, 1.2)\).

% Numerical stability is ensured by verifying positive definiteness and condition numbers below \(10^8\), with ill-conditioned cases regenerated using an incremented random seed. By emphasizing the \emph{standard normal distribution} for matrix entries and \emph{uniform distributions} for parameters like \(\beta\), sparsity scaling, and \(\mathbf{b}\), this dataset provides diverse, well-conditioned systems tailored for evaluating the RL-based precision selection routine.

The empirical results are shown in \tablename~\ref{tab:pl_fp64_metrics_sum1}. The RL-based mixed-precision CG solver exhibits mean relative errors of $6.81 \times 10^{-4}$ under cost setting $C_1$  and $6.93 \times 10^{-4}$ under $C_2$, compared to $4.21 \times 10^{-4}$ for the fp64-CG solver. While RL-CG errors are approximately 1.6 times higher, they remain within acceptable bounds ($< 10^{-3}$), with standard deviations ($1.38 \times 10^{-3}$ for $C_1$, $1.23 \times 10^{-3}$ for $C_2$) exceeding fp64-CG’s ($8.39 \times 10^{-4}$), indicating greater variability due to mixed-precision operations. The maximum errors for RL-CG ($1.15 \times 10^{-2}$ for $C_1$, $8.05 \times 10^{-3}$ for $C_2$) suggest occasional instability compared to fp64-CG ($6.70 \times 10^{-3}$), but percentile ranges (RL-CG: $1.85 \times 10^{-4}$ to $7.11 \times 10^{-4}$; fp64-CG: $3.00 \times 10^{-8}$ to $5.27 \times 10^{-4}$) confirm competitive accuracy. The marginal error increase under $C_2$ reflects the RL agent’s preference for lower precisions, driven by higher costs for \texttt{fp32} and \texttt{fp64}. 

Besides, averaged iteration counts for RL-CG is 215 under $C_1$ and 220 under $C_2$, compared to 189 for fp64-CG, indicating an 8--16\% increase due to slower convergence from lower precisions. High standard deviations (RL-CG: 394 for $C_1$, 402 for $C_2$; fp64-CG: 380) reflect variability in matrix conditioning, with all solvers reaching the maximum 1000 iterations for some cases. The 75th percentile for RL-CG (32 for $C_1$, 26 for $C_2$) exceeds fp64-CG’s (11), suggesting fp64-CG converges faster for most matrices. The increase under $C_2$ likely stems from the RL agent’s bias toward low-cost precisions (e.g., \texttt{bf16}=0.4, \texttt{fp16}=0.5), which may reduce numerical stability. Despite this, RL-CG’s iteration counts remain comparable, supporting its practical viability.

\begin{figure}[ht]
\centering
\subfigure[Cost setting $C_1$]{\includegraphics[width=6.3cm]{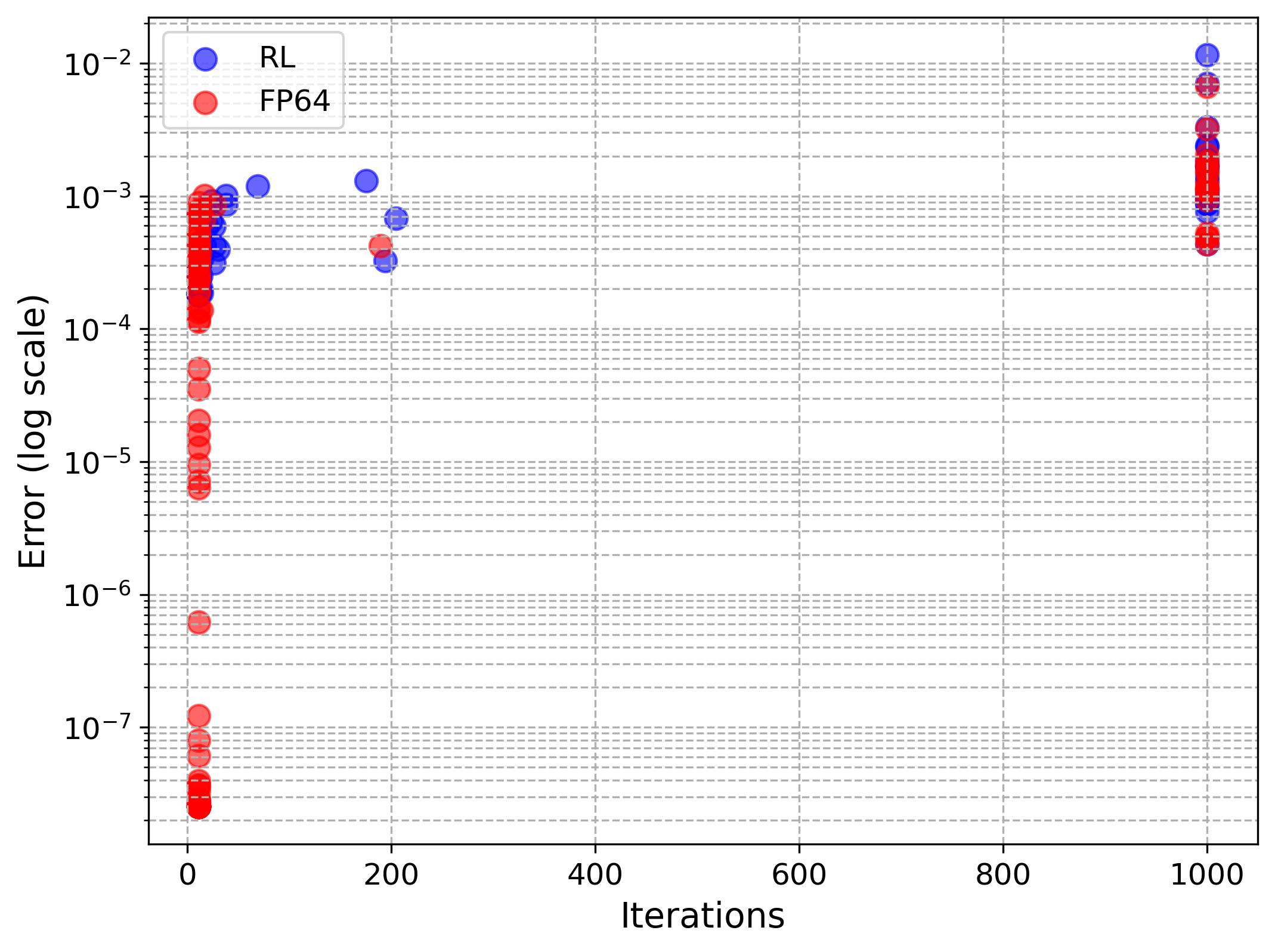}}
\subfigure[Cost setting $C_2$]{\includegraphics[width=6.3cm]{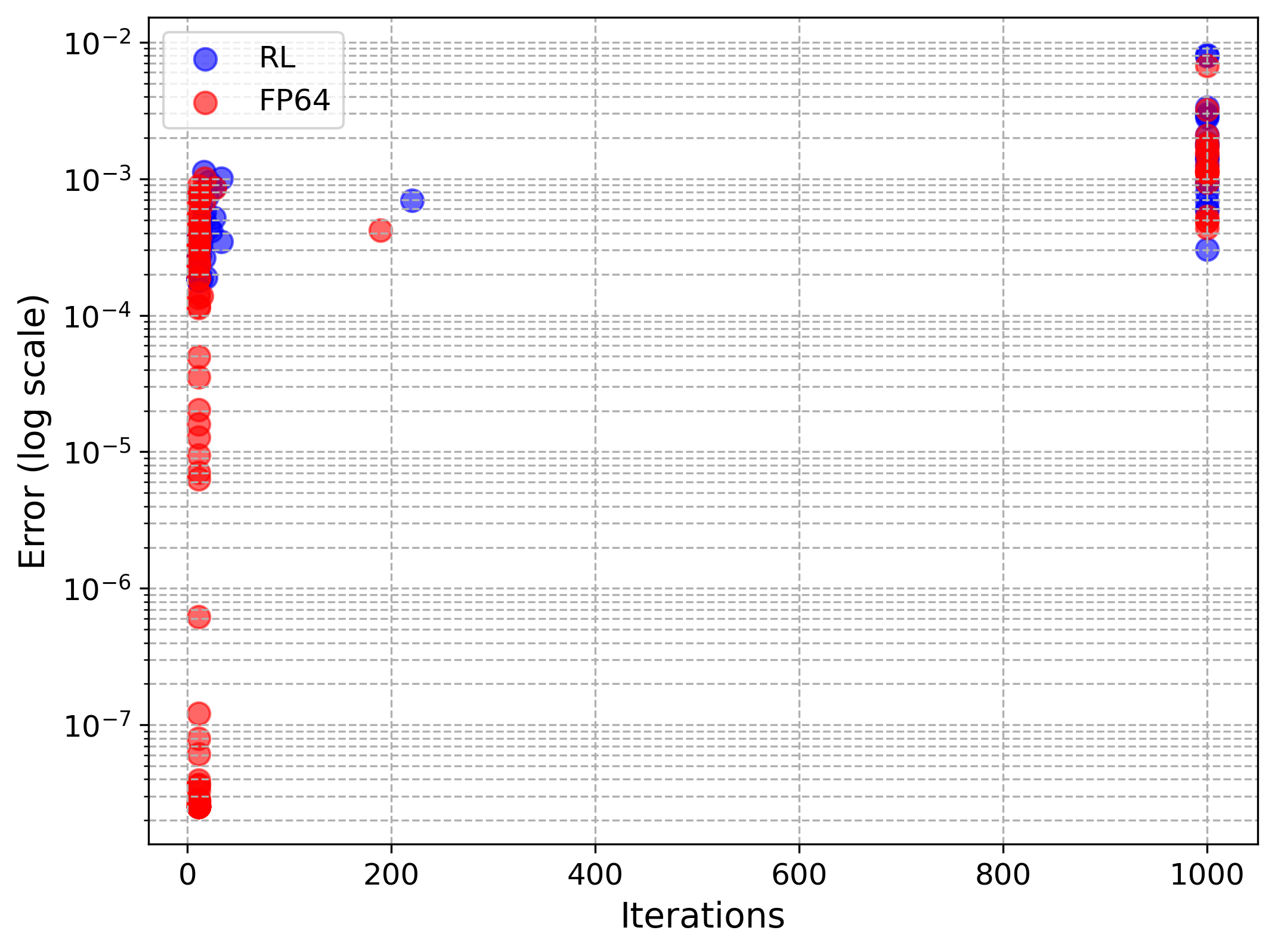}}
\caption{Comparison of RL-CG vs fp64-CG: Error vs Iterations.}\label{fig:rlfp64scatter1}
\end{figure}

As shown in \figurename~\ref{fig:rlfp64scatter1}, RL-CG and fp64-CG produce nearly overlapping results, indicating similar solution quality across both cost settings. According to \tablename~\ref{tab:average_precs} and \tablename~\ref{tab:pl_fp64_metrics_sum1}, RL-CG achieves comparable accuracy to fp64-CG while utilizing significantly more low-precision arithmetic, though with a slight increase in iteration counts. Under cost setting $C_1$, RL-CG assigns a higher proportion of high-precision operations, leading to reduced relative error and fewer iterations compared to $C_2$. Conversely, cost setting $C_2$ favors more aggressive use of lower-precision formats (e.g., fp16, bf16), yielding substantial computational savings at a modest trade-off in accuracy and convergence speed. This precision adaptation, particularly evident under $C_2$, demonstrates RL-CG’s ability to dynamically balance performance and efficiency in resource-constrained environments such as GPUs, making it a compelling solver for large-scale linear systems.

% Comparing the RL-CG and fp64 under two cost settings in \figurename~\ref{fig:rlfp64scatter1}, the two results are almost overlapped. As indicated in \figurename~\ref{tab:average_precs} and \tablename~\ref{tab:pl_fp64_metrics_sum1}, RL-CG achieve similar accuracy while performing more low precision arithmetic than fp64-CG though trigger slightly more iterations. Under $C_1$, RL-CG’s use higher proportion of higher precision operations compared to $C_2$ but the performance naturally increase as reduced interations and decreased relative error. This precision assignment, particularly pronounced under $C_2$, highlights RL-CG’s efficiency in resource-constrained settings, such as GPU-based systems. The RL-CG solver thus effectively balances accuracy and cost, with $C_2$ enhancing savings at a modest accuracy and efficiency trade-off, making it a compelling approach for large-scale linear systems.

\begin{table}[ht]
\centering
\caption{Updated statistical indices for RL-based mixed-precision CG and fp64-CG solvers across two cost settings. Metrics include relative error and iteration count. } % All values are reported to three significant digits. fp64 metrics are identical across settings and listed once.
\label{tab:pl_fp64_metrics_sum1}
\scriptsize
\setlength\tabcolsep{0.1pt} % Adjusted for readability
\renewcommand{\arraystretch}{0.8}
\begin{tabular}{l
                S[round-mode=figures, round-precision=3]
                S[round-mode=figures, round-precision=3]
                S[round-mode=figures, round-precision=3]
                S[round-mode=figures, round-precision=3]
                S[round-mode=figures, round-precision=3]
                S[round-mode=figures, round-precision=3]}
\toprule
\textbf{Metric} & \textbf{Mean} & \textbf{Std} & \textbf{Min} & \textbf{Max} & \textbf{25\%} & \textbf{75\%} \\
\midrule
\multicolumn{7}{c}{\textbf{Cost setting $C_1$}} \\
\rowcolor{blue!5}
RL Error        & 6.80e-4 & 1.38e-3 & 1.80e-4 & 1.15e-2 & 1.85e-4 & 7.11e-4 \\
\rowcolor{blue!5}
RL Iterations   & 215     & 394     & 11.0    & 1000    & 11.0    & 42.0    \\
\midrule
\multicolumn{7}{c}{\textbf{Cost setting $C_2$}} \\
\rowcolor{blue!5}
RL Error        & 6.93e-4 & 1.23e-3 & 1.80e-4 & 8.05e-3 & 1.85e-4 & 6.93e-4 \\
\rowcolor{blue!5}
RL Iterations   & 220     & 402     & 11.0    & 1000    & 11.0    & 26.0    \\
\midrule
\multicolumn{7}{c}{\textbf{fp64 (Reference)}} \\
\rowcolor{gray!10}
fp64-CG Error      & 4.21e-4 & 8.39e-4 & 3.00e-8 & 6.70e-3 & 3.00e-8 & 5.27e-4 \\
\rowcolor{gray!10}
fp64-CG Iterations & 189     & 380     & 11.0    & 1000    & 11.0    & 11.0    \\
\bottomrule
\end{tabular}
\end{table}

\subsection{2D Poisson PDE problems}
To evaluate our reinforcement learning (RL)-based mixed-precision strategy for iterative solvers of linear systems arising from partial differential equations (PDEs), we constructed a diverse dataset of systems \(\mathbf{A} \mathbf{x} = \mathbf{b}\) by discretizing the 2D Poisson equation  
\[
-\nabla^2 u = f \quad \text{in} \ \Omega = [0, 2] \times [0, 2],
\]  
subject to Dirichlet boundary conditions \(u = g\) on \(\partial \Omega\). The domain \(\Omega\) was subdivided into randomly sampled subdomains to introduce variability in the computational grid. A uniform grid with \(n_x = n_y = 80\) interior points (\(n = 6400\)) was employed, with grid spacings  
\[
h_x = \frac{b_x - a_x}{n_x + 1}, \quad h_y = \frac{b_y - a_y}{n_y + 1}.
\]  
The Laplacian was discretized using a five-point finite difference stencil, yielding a sparse system matrix \(\mathbf{A} \in \mathbb{R}^{n \times n}\). The right-hand side vector \(\mathbf{b} \in \mathbb{R}^n\) incorporated both the source term \(f(x, y)\) and contributions from boundary conditions. Besides, diversity across datasets was introduced by varying three parameters. Subdomain boundaries \([a_x, b_x] \times [a_y, b_y]\) were sampled with \(a_x \sim \mathrm{Uniform}(0, 1.9)\), \(b_x = a_x + \mathrm{Uniform}(0.1, 2 - a_x)\), and analogously for \(a_y, b_y\). Boundary conditions on each edge were randomly assigned as constant, linear, or sinusoidal functions, with parameters drawn from uniform distributions. Source terms were chosen independently as zero, sinusoidal, or polynomial functions, each with equal probability. This variation ensured that both training and testing sets covered a broad range of PDE configurations, supporting robust evaluation of the RL-based precision selection method.

\begin{table}[ht]
\centering
\caption{Statistical indices for RL-based mixed-precision CG and fp64-CG solvers across two datasets. Metrics include relative error and iteration count. } % All values are reported to three significant digits. fp64 metrics are identical across settings and listed once.
\label{tab:pl_fp64_metrics_sum2}
\scriptsize
\setlength\tabcolsep{0.1pt} % Adjusted for readability
\renewcommand{\arraystretch}{0.8}
\begin{tabular}{l
                S[round-mode=figures, round-precision=3]
                S[round-mode=figures, round-precision=3]
                S[round-mode=figures, round-precision=3]
                S[round-mode=figures, round-precision=3]
                S[round-mode=figures, round-precision=3]
                S[round-mode=figures, round-precision=3]}
\toprule
\textbf{Metric} & \textbf{Mean} & \textbf{Std} & \textbf{Min} & \textbf{Max} & \textbf{25\%} & \textbf{75\%} \\
\midrule
\multicolumn{7}{c}{\textbf{Cost setting $C_1$}} \\
\rowcolor{blue!5}
RL Error        & 2.51e-5 & 2.12e-5 & 1.00e-8 & 9.42e-5 & 6.72e-6 & 3.98e-5 \\
\rowcolor{blue!5}
RL Iterations   & 11.0    & 0.0     & 11.0    & 11.0    & 11.0    & 11.0    \\
\midrule
\multicolumn{7}{c}{\textbf{Cost setting $C_2$}} \\
\rowcolor{blue!5}
RL Error        & 7.32e-4 & 5.77e-4 & 1.49e-4 & 2.52e-3 & 3.11e-4 & 9.72e-4 \\
\rowcolor{blue!5}
RL Iterations   & 10.4    & 1.36    & 7.00    & 11.0    & 11.0    & 11.0    \\
\midrule
\multicolumn{7}{c}{\textbf{fp64 (Reference)}} \\
\rowcolor{gray!10}
fp64-CG Error      & 2.51e-5 & 2.12e-5 & 1.00e-8 & 9.42e-5 & 6.72e-6 & 3.98e-5 \\
\rowcolor{gray!10}
fp64-CG Iterations & 11.0    & 0.0     & 11.0    & 11.0    & 11.0    & 11.0    \\
\bottomrule
\end{tabular}
\end{table}

\begin{figure}[ht]
\centering
\subfigure[Cost setting $C_1$]{\includegraphics[width=6.3cm]{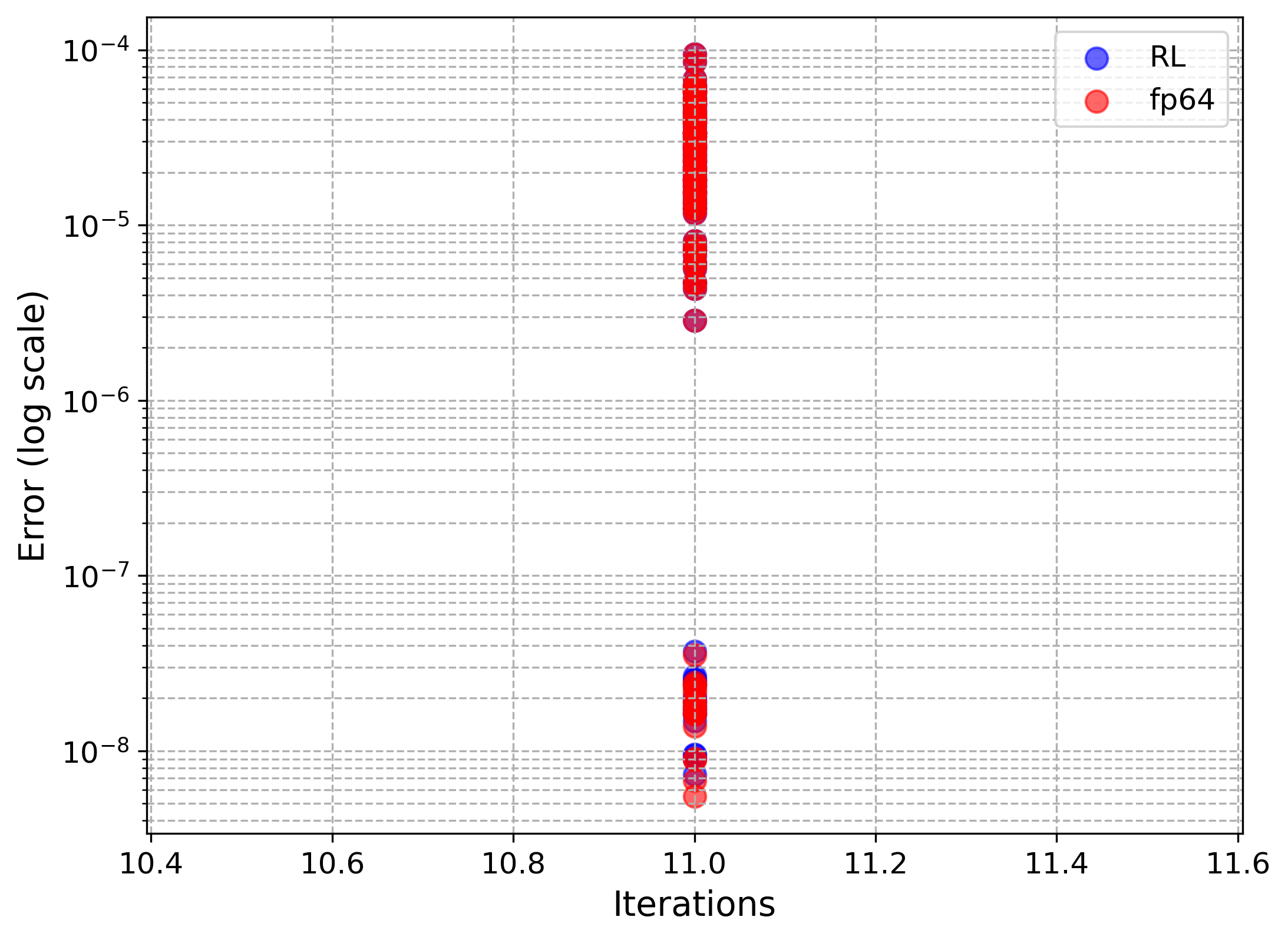}}
\subfigure[Cost setting $C_2$]{\includegraphics[width=6.2cm]{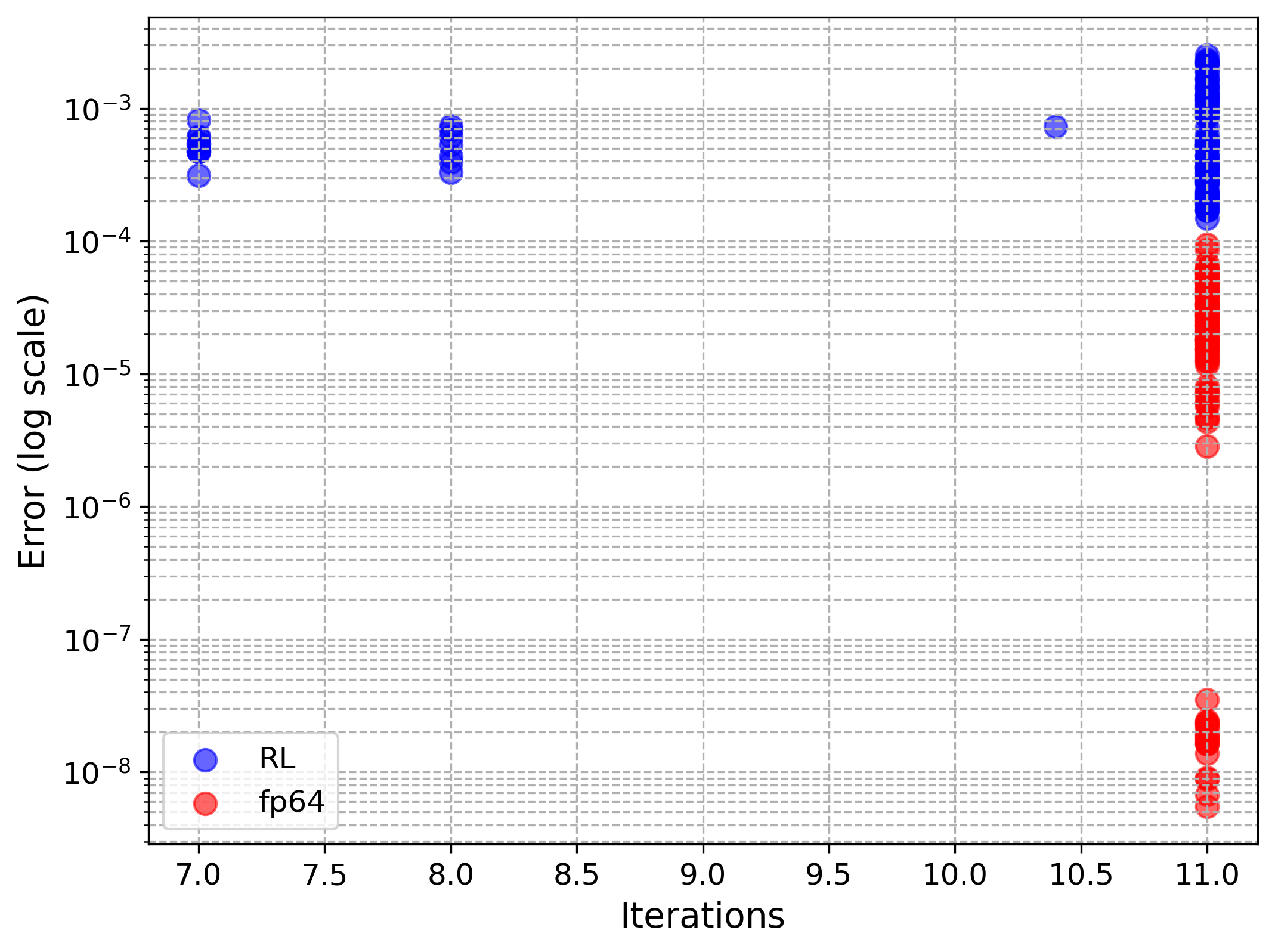}}
\caption{Comparison of RL-CG vs fp64-CG: Error vs Iterations.}\label{fig:rlfp64scatter2}
\end{figure}

In terms of the empirical results shown in \tablename~\ref{tab:pl_fp64_metrics_sum1} and \figurename~\ref{fig:rlfp64scatter2}, RL-CG solver demonstrates exceptional accuracy for 2D Poisson problem under cost setting $C_1$, achieving a mean relative error of $2.51 \times 10^{-5}$, identical to the fp64-CG solver, with matching statistical indices (standard deviation $2.12 \times 10^{-5}$, minimum $1.00 \times 10^{-8}$, maximum $9.42 \times 10^{-5}$, percentiles $6.72 \times 10^{-6}$ to $3.98 \times 10^{-5}$). This equivalence suggests that the RL agent, operating under $C_1$, selects high-precision formats (e.g., \texttt{fp64}, \texttt{fp32}) to ensure numerical stability for the dataset’s well-conditioned matrices. Conversely, under $C_2$, the mean error increases to $7.32 \times 10^{-4}$ (standard deviation $5.77 \times 10^{-4}$, range $1.49 \times 10^{-4}$ to $2.52 \times 10^{-3}$), approximately 29 times higher than fp64-CG, reflecting the RL agent’s preference for low-cost precisions (\texttt{bf16}, \texttt{fp16}) driven by $C_2$’s higher costs for \texttt{fp32} and \texttt{fp64}. Despite this, RL-CG errors under $C_2$ remain within practical tolerances ($< 10^{-3}$), indicating viability for applications prioritizing computational efficiency.

Further, the averaged iteration counts for RL-CG under $C_1$ are uniformly 11.0, aligning with fp64-CG across all metrics (standard deviation 0.0, range 11.0 to 11.0), consistent with the minimum iteration constraint and the dataset’s highly well-conditioned matrices, which facilitate rapid convergence. Under $C_2$, RL-CG averages 10.4 iterations (standard deviation 1.36, minimum 7.0, maximum 11.0), slightly lower than fp64-CG’s 11.0, likely due to early termination from numerical instabilities (e.g., invalid scalar computations) rather than improved convergence, as evidenced by the elevated error. The tight iteration range across both settings underscores the dataset’s homogeneity, limiting the RL agent’s ability to optimize convergence but highlighting its precision selection’s impact on accuracy and cost trade-offs.

% Computational cost, inferred from precision logs, is lower for RL-CG than fp64-CG’s fixed 8.0 per iteration (4 $\times$ 2.0). Under $C_1$, RL-CG’s fp64-like accuracy implies frequent use of \texttt{fp64}, with occasional \texttt{bf16} (0.6) or \texttt{fp16} (0.8), yielding a per-iteration cost of $\sim$7.0--8.0. Under $C_2$, the RL agent’s bias toward \texttt{bf16} and \texttt{fp16} reduces this to $\sim$2.8, offering significant savings despite minor iteration variability. This cost advantage, particularly under $C_2$, positions RL-CG as an efficient solution for resource-constrained environments, such as GPU-based systems, though $C_2$’s accuracy trade-off necessitates careful parameter tuning. The RL-CG solver thus excels in balancing cost and accuracy, with $C_1$ matching fp64-CG’s precision and $C_2$ prioritizing cost efficiency for well-conditioned linear systems.

\section{Discussion}
\label{sec:discussion}

% Our RL framework, powered by Q-learning, redefines precision optimization in iterative solvers. Below, we delve into its strengths, dynamics, challenges, comparisons, and future potential, illuminating its transformative impact.

Our RL framework, powered by Q-learning, redefines precision optimization in iterative solvers. Q-learning’s model-free nature eliminates the need to model CG’s complex dynamics, learning directly from solver feedback \cite{sutton2018reinforcement}. Its tabular approach suits our finite MDP ($|\mathcal{S}| = b \cdot r$, $|\mathcal{A}_j| = |\mathcal{P}|$), ensuring convergence without neural network overhead \cite{watkins1992q}. Separate Q-tables per operation scale linearly ($O(m |\mathcal{P}|)$), unlike joint tables ($O(|\mathcal{P}|^m)$). This modularity enables precise control—e.g., assigning $\texttt{fp16}$ to matrix-vector products early, saving costs, while reserving $\texttt{fp64}$ for inner products near convergence. Q-learning’s adaptability to residual changes makes it robust across matrix properties, from well-conditioned to ill-conditioned systems.

% \subsection{Precision Optimization Dynamics}
% Q-learning learns context-aware precision schedules, a leap beyond static rules. Early iterations, with large $\rho_k$, favor $\texttt{fp16}$ or $\texttt{fp32}$, reducing costs when errors are tolerable—e.g., a matrix-vector product might use $\texttt{fp16}$ for a sparse $A$. As $\rho_k \to \tau$, Lemma \ref{lem:policy_monotonicity} shows RL shifts to $\texttt{fp64}$, ensuring stability (Proposition \ref{prop:policy_stability}). For instance, in a physics simulation with a stiff matrix, RL might learn to use $\texttt{fp32}$ for preconditioner solves mid-iteration, switching to $\texttt{fp64}$ when $\rho_k < 10^{-6}$, balancing speed and accuracy. This dynamic adaptation outperforms fixed precision, minimizing runtime while meeting $\tau$.

\section{Conclusion}
\label{sec:conclusion}
In this work, we introduce a transformative reinforcement learning framework for precision selection in conjugate gradient (CG) solvers via elegantly merging Q-learning’s adaptability with the stringent demands of numerical precision. By dynamically learning precision schedules, our framework achieves a level of efficiency and flexibility that existing precision tuning methods cannot rival, thereby unlocking unprecedented computational efficiencies. The RL-CG solver effectively balances accuracy and cost, with  customized cost settings enhancing savings at a modest accuracy and efficiency trade-off, making it a compelling approach for large-scale linear systems. Since the MDP modeling does not depend on matrix size, this adaptability allows our method to generalize across matrices of varying dimensions, enabling training on smaller datasets while efficiently inferring solutions for larger, arbitrarily scaled systems. This critical advantage mitigates the dependency on extensive data, as validated through our experiments, which demonstrate robust generalization across diverse data source. In future work, we will investigate deep reinforcement learning, non-SPD systems, and hardware integration, paving the way for a new era of approximate computing.

%\section{Acknowledgement}
%This work was supported by the France 2030 NumPEx Exa-MA (ANR-22-EXNU-0002) project managed by the French National Research Agency (ANR). I thank Xiaobo Liu for interesting discussion and suggestions. 

\bibliographystyle{abbrv}
\bibliography{references}

\begin{thebibliography}{10}

\bibitem{osti_1814677}
A.~Abdelfattah, H.~Anzt, A.~Ayala, E.~Boman, E.~Carson, S.~Cayrols, T.~Cojean, J.~Dongarra, R.~Falgout, M.~Gates, et~al.
\newblock Advances in mixed precision algorithms: 2021 edition.
\newblock Technical report, Lawrence Livermore National Lab. (LLNL), Livermore, CA (United States), 2021.

\bibitem{abdelfattah2021survey}
A.~Abdelfattah, H.~Anzt, E.~G. Boman, E.~Carson, T.~Cojean, J.~Dongarra, A.~Fox, M.~Gates, N.~J. Higham, X.~S. Li, P.~Luszczek, S.~Pranesh, and S.~Tomov.
\newblock A survey of numerical linear algebra methods utilizing mixed-precision arithmetic.
\newblock {\em The International Journal of High Performance Computing Applications}, 35(4):344--369, 2021.

\bibitem{anzt2021mixed}
H.~Anzt, M.~Baboulin, J.~Dongarra, G.~Flegar, N.~J. Higham, P.~Luszczek, and S.~Tomov.
\newblock Mixed-precision methods for linear solvers in scientific computing: Theory and practice.
\newblock {\em Computer Science Review}, 40:100430, 2021.

\bibitem{anzt2020adaptive}
H.~Anzt, T.~Cojean, G.~Flegar, S.~Tomov, and J.~Dongarra.
\newblock Adaptive precision in iterative methods for sparse linear systems.
\newblock {\em Parallel Computing}, 93:102602, 2020.

\bibitem{baboulin2009accelerating}
M.~Baboulin, A.~Buttari, J.~Dongarra, J.~Kurzak, J.~Langou, J.~Langou, P.~Luszczek, and S.~Tomov.
\newblock Accelerating scientific computations with mixed precision algorithms.
\newblock {\em Computer Physics Communications}, 180(12):2526--2533, 2009.

\bibitem{bottou2018optimization}
L.~Bottou, F.~E. Curtis, and J.~Nocedal.
\newblock Optimization methods for large-scale machine learning.
\newblock {\em SIAM Review}, 60(2):223--311, 2018.

\bibitem{carson2025}
E.~Carson and X.~Chen.
\newblock \texttt{Pychop}: Emulating low-precision arithmetic in numerical methods and neural networks, 2025.

\bibitem{doi:10.1137/17M1122918}
E.~Carson and N.~J. Higham.
\newblock A new analysis of iterative refinement and its application to accurate solution of ill-conditioned sparse linear systems.
\newblock {\em SIAM Journal on Scientific Computing}, 39(6):A2834--A2856, 2017.

\bibitem{carson2021mixed}
E.~Carson and N.~J. Higham.
\newblock Accelerating the solution of linear systems by iterative refinement in three precisions.
\newblock {\em SIAM Journal on Scientific Computing}, 40(2):A817--A847, 2018.

\bibitem{carson2019accelerating}
E.~Carson, N.~J. Higham, and S.~Pranesh.
\newblock Accelerating iterative refinement for linear systems using {GMRES}-based preconditioners.
\newblock {\em SIAM Journal on Scientific Computing}, 41(5):A3333--A3355, 2019.

\bibitem{doi:10.1137/20M1316822}
E.~Carson, N.~J. Higham, and S.~Pranesh.
\newblock Three-precision gmres-based iterative refinement for least squares problems.
\newblock {\em SIAM Journal on Scientific Computing}, 42(6):A4063--A4083, 2020.

\bibitem{davis2006direct}
T.~A. Davis.
\newblock {\em Direct Methods for Sparse Linear Systems}.
\newblock Society for Industrial and Applied Mathematics, 2006.

\bibitem{deroos2017krylov}
F.~de~Roos and P.~Hennig.
\newblock Krylov subspace recycling for fast iterative least-squares in machine learning, 2017.

\bibitem{demmel1997applied}
J.~W. Demmel.
\newblock {\em Applied Numerical Linear Algebra}.
\newblock Society for Industrial and Applied Mathematics, 1997.

\bibitem{dennis2010computational}
J.~M. Dennis, J.~Edwards, K.~J. Evans, R.~Loy, S.~A. Mickelson, D.~Stammer, M.~A. Taylor, M.~Vertenstein, and P.~H. Worley.
\newblock Computational challenges in high-resolution climate modeling.
\newblock {\em Computing in Science \& Engineering}, 12(5):18--25, 2010.

\bibitem{graillat2018promise}
S.~Graillat, F.~J{\'e}z{\'e}quel, R.~Picot, F.~F{\'e}votte, and B.~Lathuili{\`e}re.
\newblock Auto-tuning for floating-point precision with discrete stochastic arithmetic.
\newblock {\em Journal of Computational Science}, 36:101017, 2019.

\bibitem{hestenes1952methods}
M.~R. Hestenes and E.~Stiefel.
\newblock Methods of conjugate gradients for solving linear systems.
\newblock {\em Journal of Research of the National Bureau of Standards}, 49(6):409--436, 1952.

\bibitem{higham2002accuracy}
N.~J. Higham.
\newblock {\em Accuracy and Stability of Numerical Algorithms}.
\newblock Society for Industrial and Applied Mathematics, 2nd edition, 2002.

\bibitem{higham2019mixed}
N.~J. Higham and T.~Mary.
\newblock Mixed precision numerical linear algebra: A survey.
\newblock {\em Numerical Linear Algebra with Applications}, 26(5):e2263, 2019.

\bibitem{10.1145/3649329.3656231}
E.~Kwon, M.~Zhou, W.~Xu, T.~Rosing, and S.~Kang.
\newblock Rl-ptq: Rl-based mixed precision quantization for hybrid vision transformers.
\newblock In {\em Proceedings of the 61st ACM/IEEE Design Automation Conference}, DAC '24, New York, NY, USA, 2024. Association for Computing Machinery.

\bibitem{luo2024neural}
J.~Luo, J.~Wang, H.~Wang, H.~Dong, Z.~Geng, H.~Chen, and Y.~Kuang.
\newblock Neural krylov iteration for accelerating linear system solving.
\newblock In {\em Advances in Neural Information Processing Systems}, volume~37. Curran Associates, Inc., 2024.

\bibitem{markidis2018nvidia}
S.~Markidis, S.~W.~D. Chien, E.~Laure, I.~B. Peng, and J.~S. Vetter.
\newblock Nvidia tensor core programmability, performance, and precision.
\newblock {\em The International Journal of High Performance Computing Applications}, 34(1):45--58, 2018.

\bibitem{meijerink1977iterative}
J.~A. Meijerink and H.~A. van~der Vorst.
\newblock An iterative solution method for linear systems of which the coefficient matrix is a symmetric {M}-matrix.
\newblock {\em Mathematics of Computation}, 31(137):148--162, 1977.

\bibitem{mnih2015human}
V.~Mnih, K.~Kavukcuoglu, D.~Silver, A.~A. Rusu, J.~Veness, M.~G. Bellemare, A.~Graves, M.~Riedmiller, A.~K. Fidjeland, G.~Ostrovski, S.~Petersen, C.~Beattie, A.~Sadik, I.~Antonoglou, H.~King, D.~Kumaran, D.~Wierstra, S.~Legg, and D.~Hassabis.
\newblock Human-level control through deep reinforcement learning.
\newblock {\em Nature}, 518(7540):529--533, 2015.

\bibitem{carson2022mixed}
E.~Oktay and E.~Carson.
\newblock Mixed precision {GMRES}-based iterative refinement with recycling, 2022.

\bibitem{NEURIPS2019_9015}
A.~Paszke, S.~Gross, F.~Massa, A.~Lerer, J.~Bradbury, G.~Chanan, T.~Killeen, Z.~Lin, N.~Gimelshein, L.~Antiga, A.~Desmaison, A.~Kopf, E.~Yang, Z.~DeVito, M.~Raison, A.~Tejani, S.~Chilamkurthy, B.~Steiner, L.~Fang, J.~Bai, and S.~Chintala.
\newblock {PyTorch}: An imperative style, high-performance deep learning library.
\newblock In {\em Advances in Neural Information Processing Systems}, volume~32, pages 8024--8035. Curran Associates, Inc., 2019.

\bibitem{rubio2013precimonious}
C.~Rubio-González, C.~Nguyen, H.~D. Nguyen, J.~Demmel, W.~Kahan, K.~Sen, D.~H. Stern, and D.~H. Bailey.
\newblock Precimonious: Tuning assistant for floating-point precision.
\newblock {\em Proceedings of the International Conference for High Performance Computing, Networking, Storage and Analysis}, pages 1--12, 2013.

\bibitem{saad2003iterative}
Y.~Saad.
\newblock {\em Iterative Methods for Sparse Linear Systems}.
\newblock Society for Industrial and Applied Mathematics, 2nd edition, 2003.

\bibitem{sutton2018reinforcement}
R.~S. Sutton and A.~G. Barto.
\newblock {\em Reinforcement Learning: An Introduction}.
\newblock MIT Press, 2nd edition, 2018.

\bibitem{2020SciPy-NMeth}
P.~Virtanen, R.~Gommers, T.~E. Oliphant, M.~Haberland, T.~Reddy, D.~Cournapeau, E.~Burovski, P.~Peterson, W.~Weckesser, J.~Bright, S.~J. van~der Walt, M.~Brett, J.~Wilson, K.~J. Millman, N.~Mayorov, A.~R.~J. Nelson, E.~Jones, R.~Kern, E.~Larson, C.~J. Carey, {\.I}.~Polat, Y.~Feng, E.~W. Moore, J.~VanderPlas, D.~Laxalde, J.~Perktold, R.~Cimrman, I.~Henriksen, E.~A. Quintero, C.~R. Harris, A.~M. Archibald, A.~H. Ribeiro, F.~Pedregosa, P.~van Mulbregt, and {SciPy 1.0 Contributors}.
\newblock {SciPy} 1.0: Fundamental algorithms for scientific computing in python.
\newblock {\em Nature Methods}, 17(3):261--272, 2020.

\bibitem{10566890}
Y.~Wang, S.~Guo, J.~Guo, Y.~Zhang, W.~Zhang, Q.~Zheng, and J.~Zhang.
\newblock Data quality-aware mixed-precision quantization via hybrid reinforcement learning.
\newblock {\em IEEE Transactions on Neural Networks and Learning Systems}, pages 1--14, 2024.

\bibitem{watkins1992q}
C.~J. C.~H. Watkins and P.~Dayan.
\newblock Q-learning.
\newblock {\em Machine Learning}, 8(3-4):279--292, 1992.

\bibitem{10.1007/3-540-48166-4_16}
A.~Zeller.
\newblock Yesterday, my program worked. today, it does not. why?
\newblock In O.~Nierstrasz and M.~Lemoine, editors, {\em Software Engineering --- ESEC/FSE '99}, pages 253--267, Berlin, Heidelberg, 1999. Springer.

\bibitem{zienkiewicz2005finite}
O.~C. Zienkiewicz, R.~L. Taylor, and J.~Z. Zhu.
\newblock {\em The Finite Element Method: Its Basis and Fundamentals}.
\newblock Elsevier Butterworth-Heinemann, 6th edition, 2005.

\end{thebibliography}

\end{document}